\PassOptionsToPackage{table}{xcolor}
\documentclass[runningheads]{llncs}

\usepackage{eccv}

\usepackage{eccvabbrv}

\usepackage{multirow}

\definecolor{oursrow}{RGB}{238,244,250}
\definecolor{secondbg}{RGB}{245,245,245}

\newcommand{\bestval}[1]{\textbf{#1}}
\newcommand{\secondval}[1]{\cellcolor{secondbg}#1}

\usepackage{graphicx}
\graphicspath{{figures/}}

\usepackage[section]{placeins}
\usepackage{flafter}

\newcommand{\safeincludegraphics}[4][]{%
  \includegraphics[width=#2,height=#3,keepaspectratio,pagebox=cropbox,#1]{#4}}

\usepackage{booktabs}
\usepackage{amsmath}
\usepackage{algorithm}
\usepackage{algorithmic}
\usepackage[accsupp]{axessibility}  

\DeclareRobustCommand{\corrauth}{\textsuperscript{\ensuremath{\dagger}}}

\usepackage{hyperref}

\usepackage{orcidlink}

\begin{document}

\title{DecoupleGS: Interactive 3D Gaussian Splatting for End-to-End Autonomous Driving Testing}
\titlerunning{DecoupleGS}

\author{Siying Li\inst{1,2} \and Ying Ni\inst{1,2}\corrauth{} \and Jie Sun\inst{1,2} \and Jian Sun\inst{1,2} \and Haotian Shi\inst{1,2}\corrauth{}}
\authorrunning{S.~Li et al.}

\institute{
College of Transportation, Tongji University, Shanghai 201804, China
\and
Key Laboratory of Road and Traffic Engineering, Ministry of Education, Shanghai 201804, China\\
\email{\{siyingli,ying\_ni,jie\_sun,sunjian,shihaotian95\}@tongji.edu.cn}\\
}

\maketitle
\begingroup
\renewcommand{\thefootnote}{}
\footnotetext{\corrauth{} Corresponding authors.}
\endgroup

\begin{abstract}
End-to-end (E2E) autonomous driving algorithms require rigorous closed-loop validation in simulation environments offering high visual fidelity, strong interactivity, and real-time performance. Existing approaches, from game engines to static neural rendering, inherently trade off these requirements and struggle with the dynamic scene composition essential for E2E testing. To bridge this gap, we propose a novel decoupled 3D Gaussian Splatting (3DGS) framework tailored for large-scale E2E evaluation. We fundamentally decompose scenes into a high-fidelity static background and manipulable dynamic agents using an object-centric canonical representation. To resolve resulting representational conflicts, we introduce three targeted modules: (1) asset compression via perceptual pruning and vector quantization for real-time traffic rendering; (2) map-guided geometric registration leveraging semantic topology to strictly align trajectories; and (3) proxy-based relighting transferring ambient illumination for seamless photometric integration. Extensive experiments demonstrate that DecoupleGS achieves a balanced fidelity-efficiency trade-off, improves metric and photometric consistency, and provides a practical closed-loop sensor simulation platform for E2E autonomous driving evaluation.
\keywords{Autonomous Driving \and 3D Gaussian Splatting \and End-to-End Testing \and Sensor Simulation}
\end{abstract}

\section{Introduction}
\label{sec:intro}

Autonomous driving (AD) is rapidly transitioning from classical modular pipelines to End-to-End (E2E) learning paradigms that map raw sensor observations directly to control commands \cite{codevilla2018end,tampuu2020survey,chen2023end}. While E2E algorithms reduce error accumulation \cite{hu2023planning,jiang2023vad,chitta2022transfuser}, their safety and driving policies must be rigorously optimized across a broad, demanding set of scenarios. Since real-world data collection for such safety-critical scenarios is prohibitively costly and hazardous, high-fidelity virtual simulation has become an indispensable data engine for the continuous testing and optimization of E2E algorithms.

\begin{figure}[!t]
  \centering
  \safeincludegraphics{\linewidth}{0.42\textheight}{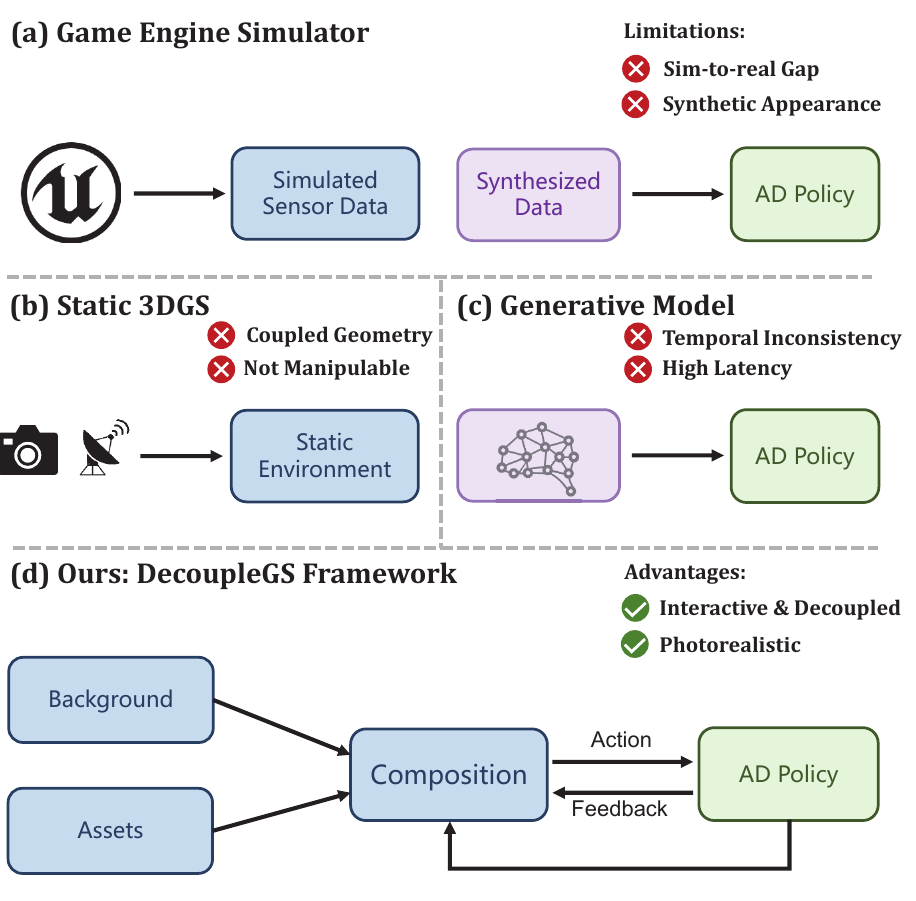}
  \caption{\textbf{Comparison of autonomous driving simulation paradigms.} Traditional approaches face distinct limitations: (a) Game engine simulators suffer from sim-to-real gaps and synthetic appearances; (b) Generative models struggle with temporal inconsistencies and high inference latency; (c) Static 3DGS representations couple geometry and illumination, making dynamic manipulation impossible. In contrast, (d) our DecoupleGS framework introduces a modular composition of backgrounds and compressed assets, uniquely enabling an interactive, photorealistic, and closed-loop testing environment for AD policies.}
  \label{fig:teaser}
\end{figure}

For E2E testing, a practical closed-loop simulation platform must satisfy three key requirements: high visual and geometric fidelity, strong multi-agent interactivity, and sufficient real-time throughput \cite{dosovitskiy2017carla,shah2018airsim,rong2020lgsvl,li2022metadrive,behrisch2011sumo}. However, existing sensor simulation approaches inherently trade off these requirements. Traditional game engines suffer from sim-to-real domain gaps, generative 2D models lack strict multi-view 3D consistency for spatial-temporal control loops, and static neural rendering methods lack the modularity to dynamically manipulate traffic flows.

Consequently, current methodologies cannot meet the integrated demands of a large-scale, rigorous testing pipeline. This gap stems from representational conflicts when seamlessly separating persistent static infrastructure from dynamic traffic participants. Merging these distinct entities introduces severe bottlenecks: storing and rendering multiple high-fidelity vehicles exhausts memory (efficiency conflict); unstructured neural coordinates fail to align with semantic metric maps (geometric conflict); and entangled environment lighting prevents dynamic assets from inheriting consistent illumination (photometric conflict).

To resolve these challenges, we propose DecoupleGS, a novel decoupled 3D Gaussian Splatting (3DGS) framework. The core task of this paper is to provide a comprehensive sensor simulation framework for E2E algorithms, achieving interactive, high-fidelity closed-loop evaluation that enables rigorous safety validation and dynamic multi-agent testing at scale. As compared in Fig. \ref{fig:teaser}, while existing simulation paradigms suffer from the sim-to-real gap, temporal inconsistency, or coupled geometry, DecoupleGS uniquely bridges these gaps by organically composing persistent backgrounds and manipulable assets. We utilize an object-centric canonical representation to decompose the scene into a persistent high-fidelity static background and independent, compact 3DGS dynamic agents. Our pipeline integrates three targeted modules: an asset compression module utilizing perceptual redundancy pruning and vector quantization to mitigate memory bottlenecks; a map-guided geometric registration module leveraging semantic topology to strictly align agent trajectories; and a proxy-based relighting mechanism transferring ambient illumination for seamless photometric integration.

The main contributions of this study are threefold. First, we present a holistic 3DGS framework that integrates neural reconstruction fidelity with the interactivity and throughput required for large-scale closed-loop E2E testing. Second, we systematically address fundamental efficiency, geometry, and photometry conflicts by developing solver-agnostic modules for vehicle compression, map-guided registration, and relighting. Third, we validate that these modules enable plug-and-play insertion of multiple vehicles at interactive rates, establishing a robust and efficient data engine for E2E algorithms.

\section{Related Work}
\label{sec:related}

\subsection{Traditional Simulators and 2D Generative Methods}
Game-engine-based simulators provide mature scene graphs, physically-based rendering primitives, and real-time rasterization pipelines, which yield strong interactivity and efficient runtime performance \cite{dosovitskiy2017carla,shah2018airsim,rong2020lgsvl,li2022metadrive,behrisch2011sumo}. Their principal limitation for testing lies in the engineering cost of producing extensive collections of high-fidelity, diverse assets, and the reliance on hand-crafted materials and approximate physics, which often preserve a non-negligible visual or sensor-level domain gap from recorded real scenes. Conversely, generative 2D image synthesis—particularly latent diffusion models—has recently achieved impressive photorealism for scene synthesis and controllable simulation \cite{rombach2022high,wang2023drivedreamer,hu2023gaia,dong2024drivearena,fang2024drivingsphere,liu2025symdrive}, but these methods are fundamentally 2D. They do not provide an intrinsic, multi-view, physically consistent 3D representation and therefore struggle with high inference latency and temporal coherence under the high frame-rate demands of control loops.

\subsection{Neural 3D Reconstruction and Rendering}
By contrast, neural 3D reconstruction and neural rendering methods—most notably NeRF variants and the recent 3DGS paradigm—reconstruct geometry and view-dependent appearance directly in a 3D coordinate frame, which reduces certain sim-to-real gaps and enables novel-view synthesis \cite{mildenhall2020nerf,barron2022mip,muller2022instant,chen2022tensorf,yu2022plenoxels,kerbl20233d,yan2024multi}. While recent frameworks advance closed-loop safety testing and dynamic scene composition \cite{zhou2024drivinggaussian,yan2024street,zhou2023hugsim,yang2023unisim,wang2024neuroncap,su2024pseudo,ge2025unraveling}, they still face bottlenecks for interactive E2E testing: many rely on pre-rendered alternatives, incorporate lighting into the reconstruction, or generally lack an efficient, map-aware modular decomposition that cleanly separates persistent infrastructure from movable traffic participants \cite{fridovich2023kplanes,tancik2022block,turki2022mega}. Consequently, naively reusing reconstructed vehicle instances across scenarios or moving them along arbitrary trajectories produces geometric and photometric inconsistencies.

\subsection{Research Gaps for Interactive Neural Scene Composition}
To address these challenges, recent attempts to make neural reconstructions interactive or to insert dynamic vehicles into neural scenes have advanced the state of the art \cite{yu2022unsupervised,wu2023mars,ost2021neural,kundu2022panoptic} but expose three concrete research gaps that directly impede closed-loop E2E evaluation.

\textbf{Efficiency and Scalability.} High-fidelity vehicle models derived from 3DGS or dense Gaussian representations typically contain a large number of primitives. Although 3DGS substantially accelerates rendering compared with early implicit NeRF-based methods, the composition of multiple detailed vehicle instances remains computationally demanding, often leading to severe frame-rate degradation when many agents co-exist in a scene. This limitation arises because most reconstruction-oriented pipelines prioritize offline visual fidelity and lack runtime-oriented compression or pruning mechanisms, which are essential for maintaining interactive performance in large-scale multi-agent simulation \cite{zhou2024drivinggaussian,yan2024street}.

\textbf{Geometric Alignment and Metric Consistency.} Neural reconstructions are typically represented in unstructured learned coordinates, whereas planning and control operate in metric map space, leading to alignment inconsistencies in closed-loop simulation. Existing object-centric methods \cite{yu2022unsupervised,ost2021neural} focus on single-object editing or require per-scene optimization and therefore do not yet provide robust, map-aware registration for long-horizon multi-agent sequences.

\textbf{Photometric Integration and Relighting.} In neural scene representations, lighting is often entangled with geometry and appearance, preventing inserted vehicles from automatically inheriting consistent ambient illumination, shadows, and environmental reflections. Although recent relighting and compositing pipelines demonstrate promising capabilities for object-level relighting and shadow synthesis \cite{zhu2024relighting}, they typically incur substantial computational overhead or assume a limited number of edited objects and fixed camera configurations. These constraints hinder their applicability to large-scale, interactive multi-vehicle simulation.

\begin{figure}[!t]
  \centering
  \safeincludegraphics{\linewidth}{0.50\textheight}{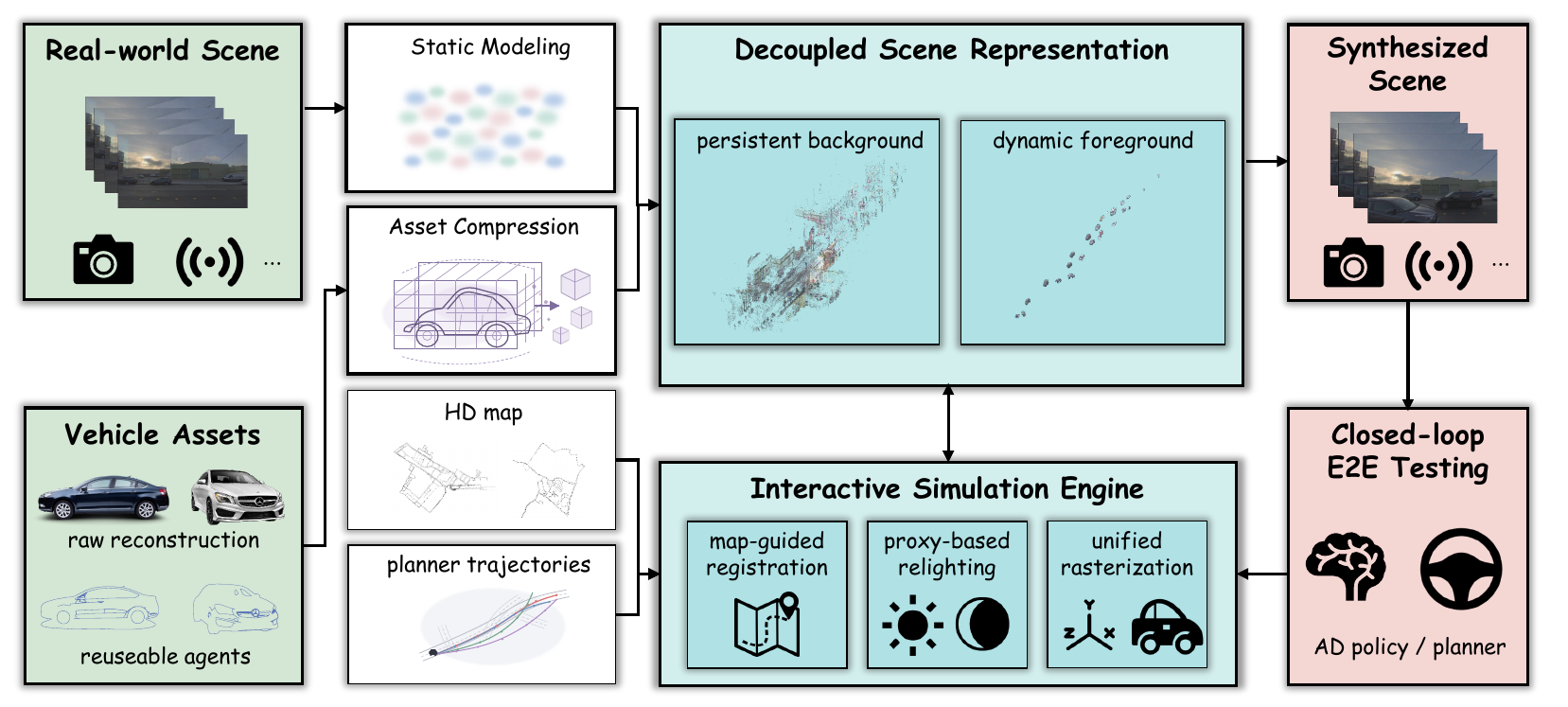}
  \caption{Overview of the DecoupleGS framework. DecoupleGS decouples persistent static backgrounds from canonical dynamic assets, and integrates asset compression, map-guided registration, proxy-based relighting, and unified rasterization to support interactive closed-loop E2E evaluation.}
  \label{fig:pipeline}
\end{figure}

\section{Method}
\label{sec:method}

To enable large-scale closed-loop E2E evaluation, we aim to compose a photorealistic yet interactive simulation where (i) the static infrastructure remains persistent across time, and (ii) a variable number of traffic agents can be instantiated, moved, relit, and rendered in response to planner-driven scene states. As illustrated in Fig. \ref{fig:pipeline}, DecoupleGS is organized around a decoupled neural scene representation and three tightly coupled functional modules. The static background and canonical dynamic assets are represented in separate coordinate spaces, while asset compression, map-guided registration, and proxy-based relighting respectively address the efficiency, geometric, and photometric conflicts that arise during dynamic scene composition. The transformed and relit primitives are finally composited through a unified rasterizer to produce multi-view observations for closed-loop E2E testing.

\subsection{Decoupling and Unified Rendering}
Applying 3DGS to interactive E2E testing is fundamentally bottlenecked by its inherent assumption of static scenes. We introduce a dual-stream architecture that represents the background and dynamic agents in separate coordinate spaces, mathematically fusing them prior to rasterization without the need for network inference.

\subsubsection{Object-Centric Canonical Decomposition.}
Formally, the global scene $\Omega$ at timestamp $t$ is defined as the spatial union of a time-invariant background field $\mathcal{S}_{bg}$ and a set of $K$ dynamic agents:
\begin{equation}
    \Omega(t) = \mathcal{S}_{bg} \cup \left( \bigcup_{k=1}^K \mathcal{T}_k(t) \circ \mathcal{V}_k \right)
\end{equation}
The background $\mathcal{S}_{bg}$ consists of Gaussian primitives anchored in the world coordinate system (WCS). Conversely, each dynamic agent $k$ is parameterized as a canonical Gaussian volume $\mathcal{V}_{k}$ within a standardized local coordinate system (LCS). Given the agent's 6-DoF rigid transform $\mathcal{T}_k(t) = [\mathbf{R}_k \mid \mathbf{t}_k] \in SE(3)$ mapping LCS to WCS, we apply the transformation to the spatial attributes (means $\mu$ and covariances $\Sigma$) of all primitives within $\mathcal{V}_k$:
\begin{equation}
    \mu_i^{(W)} = \mathbf{R}_k \mu_i^{(L)} + \mathbf{t}_k, \quad \Sigma_i^{(W)} = \mathbf{R}_k \Sigma_i^{(L)} \mathbf{R}_k^\top
    \label{eq:transform}
\end{equation}
Crucially, 3DGS encodes view-dependent appearance via Spherical Harmonics (SH). When the geometry undergoes a rotation $\mathbf{R}_k$, the intrinsic SH coefficients $c_i^{(L)}$ must be correspondingly rotated to preserve correct view-dependent reflections relative to the new orientation. We apply the Wigner D-matrix $\mathbf{D}(\mathbf{R}_k)$ to rigorously align the SH bands:
\begin{equation}
    c_i^{(W)} = \mathbf{D}(\mathbf{R}_k) c_i^{(L)}
\end{equation}
While this mathematically preserves the asset's intrinsic specularities under rigid motion, resolving the ultimate photometric mismatch with the new background illumination is deferred to our proxy-based relighting module (Sec. \ref{subsec:relighting}).

\subsubsection{Unified Rasterization and Runtime Optimization.}
To ensure physically accurate occlusion, all primitives from $\Omega(t)$ are merged into a unified aggregate list $\mathcal{P}$. For each pixel $u$, primitives overlapping $u$ are depth-sorted relative to the novel camera center and composited via continuous front-to-back $\alpha$-blending. To maintain the robust interactive frame rates required for large-scale E2E testing, we integrate several hardware-level optimizations: (a) conservative frustum culling to aggressively bound $\mathcal{P}$; (b) decoupling the sorting radix, where the static background splats are cached and incrementally reused; and (c) computing batched spatial updates exclusively for the moving canonical volumes.

\subsection{Asset Compression}
\label{subsec:asset_compression}

\begin{figure}[!t]
  \centering
  \safeincludegraphics{\linewidth}{0.50\textheight}{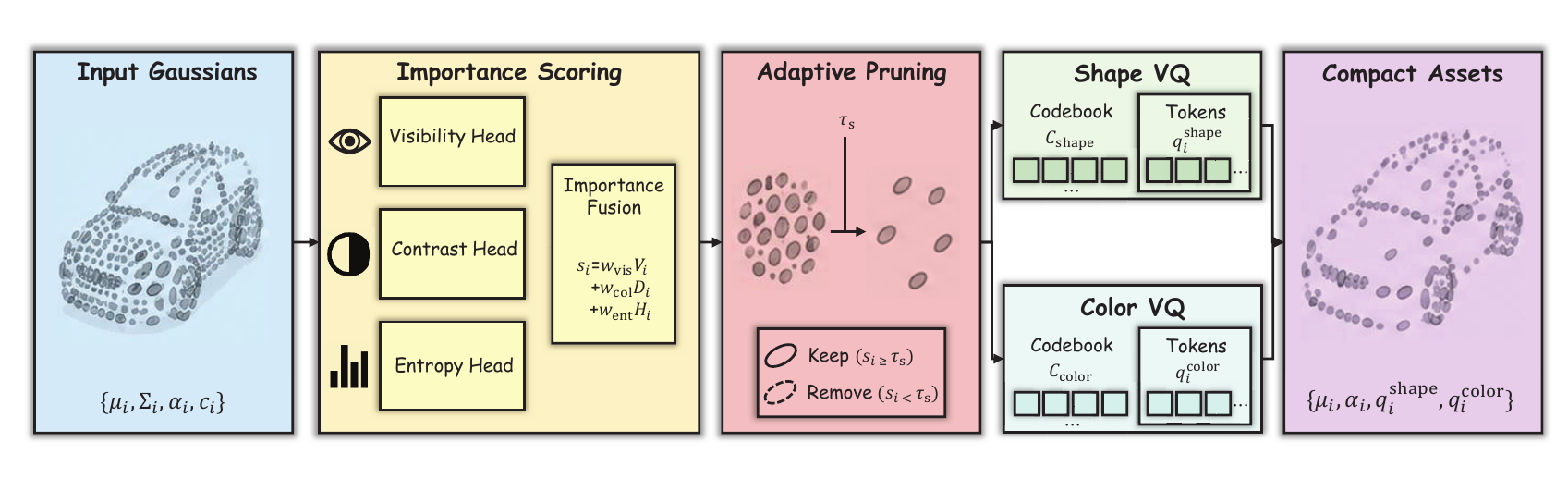}
  \caption{Illustration of the semantic-aware asset compression module. Canonical vehicle Gaussians undergo explicit importance scoring ($V_i, D_i, H_i$), adaptive pruning via threshold $\tau_s$, and high-dimensional attribute vector quantization to form compact canonical volumes.}
  \label{fig:compression}
\end{figure}

Directly rendering high-dimensional raw 3DGS assets for dense traffic is computationally prohibitive. As detailed in Fig. \ref{fig:compression}, we introduce a semantic-aware compression pipeline consisting of importance scoring, pruning, and vector quantization to produce compact canonical assets suitable for real-time multi-agent composition.

\subsubsection{Explicit Importance Scoring and Pruning.}
Instead of relying on black-box neural networks, we compute an explicit scalar importance score $s_i$ for each primitive $g_i$ that summarizes its visibility, photometric contribution, and local geometric saliency:
\begin{equation}
    s_i = w_\text{vis} V_i + w_\text{col} D_i + w_\text{ent} H_i
    \label{eq:importance}
\end{equation}
Here, the visibility estimate $V_i$ is the expected opacity contribution over sampled training views; the color contrast term $D_i$ measures perceptual distinctiveness versus the local background; and the entropy term $H_i$ captures texture complexity inside the primitive neighborhood. Primitives with $s_i$ below a strict threshold $\tau_s$ (empirically set to 0.005) are pruned. This operation significantly reduces the primitive count, concentrating computational resources on essential structural details like wheels and chassis boundaries. The weighting coefficients $w_{vis}, w_{col}$, and $w_{ent}$ are empirically determined via grid search to optimally balance geometric preservation and color fidelity, demonstrating robust generalization across different vehicle types.

\subsubsection{Attribute Vector Quantization.}
To further minimize the memory footprint, Vector Quantization (VQ) is applied to the high-dimensional attributes of retained primitives. The covariance $\Sigma_i$ and color coefficients $c_i$ are replaced by their nearest entries in small, learnable codebooks $\mathcal{C}_{shape}$ and $\mathcal{C}_{color}$:
\begin{equation}
    \Sigma_{i} \approx \mathcal{C}_\text{shape}[q_{i}^\text{shape}], \quad c_{i} \approx \mathcal{C}_\text{color}[q_{i}^\text{color}]
    \label{eq:vq}
\end{equation}
This discretization replaces heavy floating-point tensors with compact integer indices $q_i$. Specifically, we utilize attribute-specific codebooks (e.g., $K_\text{color}=1024$ for SH color coefficients and $K_\text{shape}=512$ for covariance attributes) optimized via Exponential Moving Average (EMA) during compression.

\begin{figure}[!t]
  \centering
  \safeincludegraphics{\linewidth}{0.45\textheight}{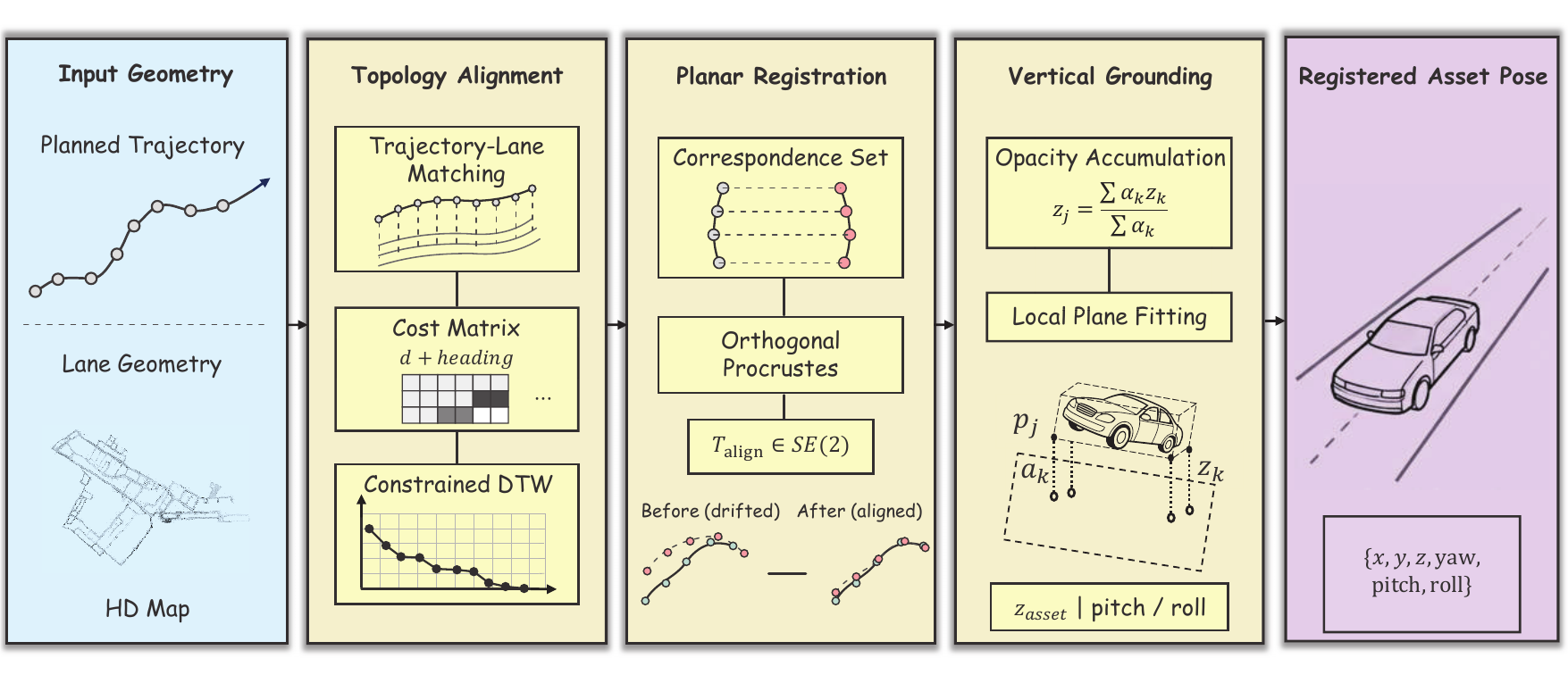}
  \caption{The map-guided geometric registration pipeline. It enforces strict metric alignment via trajectory-to-lane DTW sequence alignment, SE(2) Procrustes 2D transformation, and robust opacity-accumulated 3D vertical grounding.}
  \label{fig:registration}
\end{figure}

\subsection{Map-Guided Geometric Registration}
\label{subsec:registration}

Integrating dynamic agents requires strict geometric alignment. It is crucial to note that for the ego-vehicle evaluated in closed-loop E2E testing, its trajectory is driven entirely by the unconstrained relative pose updates from the planner to preserve maneuvers like lane changes. The sequence-level registration described below, and illustrated in Fig. \ref{fig:registration}, is specifically designed to accurately inject background traffic agents into the reconstructed shared coordinate system.

\subsubsection{Semantic Topology and Planar Alignment.}
We obtain vectorized lane centerlines $\mathcal{L}$ from dataset-provided HD maps when available, and use MapTRv2 \cite{liao2025maptrv2} to extract lane topology otherwise. Given a background agent's 2D trajectory $\mathcal{T} = \{p_t\}_{t=1}^T$ and the corresponding lane geometry, we build a cost matrix that penalizes both Euclidean distance and heading mismatch. We utilize constrained Dynamic Time Warping (DTW) to find the optimal warping path establishing trajectory-to-lane correspondences. Subsequently, an Orthogonal Procrustes analysis computes the globally optimal 2D rigid transform $T_{align} \in SE(2)$ to eliminate systemic lateral drift.

\subsubsection{Vertical Grounding via Opacity Accumulation.}
Since 3DGS lacks a continuous surface mesh, standard ray-casting fails. We employ opacity-weighted depth accumulation to ground the vehicles. For anchor points $p_j$ defined at the bottom corners of the asset's bounding box, the vertical height $z_j$ is computed as the opacity-weighted average of the intersected background gaussians within a vertical column $\mathcal{N}$:
\begin{equation}
    z_{j} = \frac{\sum_{k\in\mathcal{N}} \alpha_{k}z_{k}}{\sum_{k\in\mathcal{N}} \alpha_{k}}
    \label{eq:grounding}
\end{equation}
A local ground plane is then fitted using least-squares minimization over the anchors $\{z_j\}$. The final vertical position $z_{asset}$ and attitude (pitch and roll) of the asset are derived from the estimated plane normal, ensuring that the wheels maintain physically plausible contact with the terrain.

\subsection{Relighting}
\label{subsec:relighting}

To resolve the photometric conflict where inserted assets retain intrinsic illumination that mismatches the novel environment, we develop a lightweight, proxy-based relighting module that operates without online neural network inference.

\subsubsection{Local Probe Sampling and Linear SH Transfer.}
We treat the background SH coefficients as a dense field of light probes. For a vehicle located at $x$, we construct a local ambient descriptor $\mathcal{L}(x)$ by aggregating the SH coefficients of neighboring background gaussians $c_g$, weighted by their visibility $V_g$ and spatial distance:
\begin{equation}
    \mathcal{L}(x) = \frac{\sum_{g\in\mathcal{N}(x)} w_g c_g}{\sum_{g\in\mathcal{N}(x)} w_g}, \quad w_g = \exp\left(-\frac{||x - \mu_g||^2}{2\sigma_p^2}\right) V_g
\end{equation}
To transfer this environmental lighting, we apply a pre-calibrated linear modulation operator to the canonical vehicle's SH coefficients $c_{can}$:
\begin{equation}
    c_{out} = \mathcal{T}(\mathcal{L}(x)) \otimes c_{can} + b(\mathcal{L}(x))
    \label{eq:sh_transfer}
\end{equation}
where $\mathcal{T}(\cdot)$ and $b(\cdot)$ are efficient transformation matrices. Rather than relying on online neural network inference, these operators are derived offline by solving a least-squares regression over a synthetic dataset of canonical assets rendered under diverse pre-captured environmental probes. This analytical calibration completely avoids per-frame MLP inference while successfully rebaking the assets with the local lighting context.

\subsubsection{Proxy-based Contact Shadows.}
To anchor the object visually, we synthesize contact shadows on the estimated local ground plane. We project a parametric shadow mask $M_{shadow}(u)$ scaled by the object footprint. Crucially, the intensity of the background gaussians under this mask is modulated dynamically by the dominant light intensity $I_{dom}$ extracted from the DC component of the ambient descriptor $\mathcal{L}(x)$:
\begin{equation}
    c_\text{final}(u) = c_\text{ground}(u) \cdot \left(1 - \lambda \cdot I_\text{dom} \cdot M_\text{shadow}(u)\right)
    \label{eq:shadow}
\end{equation}
This efficient approximation ensures that shadows become appropriately faint in overcast conditions or intense under direct sunlight, completing the photometric integration seamlessly.

\section{Experiments}
\label{sec:experiments}

This section evaluates DecoupleGS from four complementary perspectives: rendering efficiency, geometric and photometric fidelity, open-loop sim-to-real consistency, and closed-loop E2E testing. We further ablate the proposed compression, registration, and relighting modules to verify how each component resolves the efficiency, geometric, and photometric conflicts in dynamic neural scene composition.

\begin{table}[!t]
  \caption{Asset compression results under different traffic densities. 
  Best values are bolded, second-best values are shaded in gray, and DecoupleGS rows are highlighted. 
  The final column summarizes whether a method consistently remains top-2 across all fidelity and efficiency metrics.}
  \label{tab:compression}
  \centering
  \small
  \setlength{\tabcolsep}{3.4pt}
  \renewcommand{\arraystretch}{1.02}
  \resizebox{\linewidth}{!}{
  \begin{tabular}{@{}clccccccc@{}}
    \toprule
    \multirow{2}{*}{Size} & \multirow{2}{*}{Methods}
    & \multicolumn{4}{c}{Fidelity}
    & \multicolumn{2}{c}{Efficiency}
    & \multirow{2}{*}{Top-2 Consist.} \\
    \cmidrule(lr){3-6} \cmidrule(lr){7-8}
    & & PSNR-Veh $\uparrow$ & PSNR-All $\uparrow$ & SSIM $\uparrow$ & LPIPS $\downarrow$ & FPS $\uparrow$ & VRAM $\downarrow$ & \\
    \midrule
    & Plenoxels~\cite{yu2022plenoxels} & 21.45 & 23.06 & 0.795 & 0.510 & 8.5 & 2.5GB & No \\
    & Vanilla 3DGS~\cite{kerbl20233d} & \bestval{28.52} & \bestval{29.41} & \bestval{0.903} & \bestval{0.243} & 55.4 & 1.2GB & No \\
    S & LightGaussian~\cite{fan2024lightgaussian} & 26.21 & 27.50 & 0.865 & 0.281 & \bestval{75.2} & \bestval{180MB} & No \\
    \rowcolor{oursrow}
    & \textbf{DecoupleGS (Ours)} & \secondval{28.10} & \secondval{29.25} & \secondval{0.898} & \secondval{0.252} & \secondval{68.5} & \secondval{850MB} & \textbf{Yes} \\
    \midrule
    & Plenoxels~\cite{yu2022plenoxels} & 20.12 & 23.08 & 0.626 & 0.463 & 6.2 & 3.1GB & No \\
    & Vanilla 3DGS~\cite{kerbl20233d} & \bestval{27.05} & \bestval{27.21} & \bestval{0.815} & \bestval{0.214} & 32.5 & 2.1GB & No \\
    M & LightGaussian~\cite{fan2024lightgaussian} & 25.14 & 26.40 & 0.781 & 0.250 & \bestval{58.6} & \bestval{240MB} & No \\
    \rowcolor{oursrow}
    & \textbf{DecoupleGS (Ours)} & \secondval{26.85} & \secondval{27.12} & \secondval{0.805} & \secondval{0.218} & \secondval{52.4} & \secondval{950MB} & \textbf{Yes} \\
    \midrule
    & Plenoxels~\cite{yu2022plenoxels} & 18.55 & 21.08 & 0.719 & 0.379 & 3.5 & 4.8GB & No \\
    & Vanilla 3DGS~\cite{kerbl20233d} & \bestval{25.50} & \bestval{25.14} & \bestval{0.785} & \bestval{0.255} & 12.1 & 3.8GB & No \\
    L & LightGaussian~\cite{fan2024lightgaussian} & 23.52 & 24.10 & 0.720 & 0.295 & \bestval{45.0} & \bestval{320MB} & No \\
    \rowcolor{oursrow}
    & \textbf{DecoupleGS (Ours)} & \secondval{25.24} & \secondval{25.05} & \secondval{0.772} & \secondval{0.260} & \secondval{38.5} & \secondval{1.1GB} & \textbf{Yes} \\
    \bottomrule
  \end{tabular}
  }
\end{table}

\subsection{Experimental Setup}

\textbf{Datasets and assets.}
We utilize nuScenes \cite{caesar2020nuscenes} and PandaSet \cite{xiao2021pandaset} for diverse background scenarios, and source manipulable canonical vehicles from 3DRealCar \cite{du20253drealcar}. We curate 15 dynamic clips from nuScenes and 10 sequences from PandaSet, covering challenging illumination conditions such as dusk, night, and overcast scenes. Dynamic objects are removed using Segment Anything Model (SAM) \cite{kirillov2023segment} masks with morphological dilation to suppress boundary artifacts. We sample 20 vehicle assets from 3DRealCar, covering sedans, SUVs, and large commercial vehicles.

\textbf{Baselines and metrics.}
For rendering efficiency and compression, we compare against Plenoxels \cite{yu2022plenoxels}, Vanilla 3DGS \cite{kerbl20233d}, and LightGaussian \cite{fan2024lightgaussian}. For simulator-level fidelity and closed-loop testing, we compare with HUGSIM \cite{zhou2023hugsim}, OASim \cite{yan2024oasim}, and RealEngine \cite{jiang2025realengine}. Downstream E2E policy testing is conducted with UniAD \cite{hu2023planning} and VAD \cite{jiang2023vad}. Rendering fidelity is measured by PSNR-Veh, PSNR-All, SSIM, and LPIPS, while runtime performance is evaluated using FPS and VRAM. Geometric consistency is quantified by Trajectory Average Displacement Error (ADE) and Ground Penetration Rate (GPR). Photometric realism is evaluated using Masked PSNR, Peak Intensity Error (PIE), and Peak Angular Error (PAE). Open-loop planner consistency is measured by mADE and minTTC, while closed-loop E2E testing adopts Driving Score (DS), Success Rate (SR), Route Completion (RC), and minTTC.

\begin{figure}[!t]
  \centering
  \includegraphics[width=\linewidth]{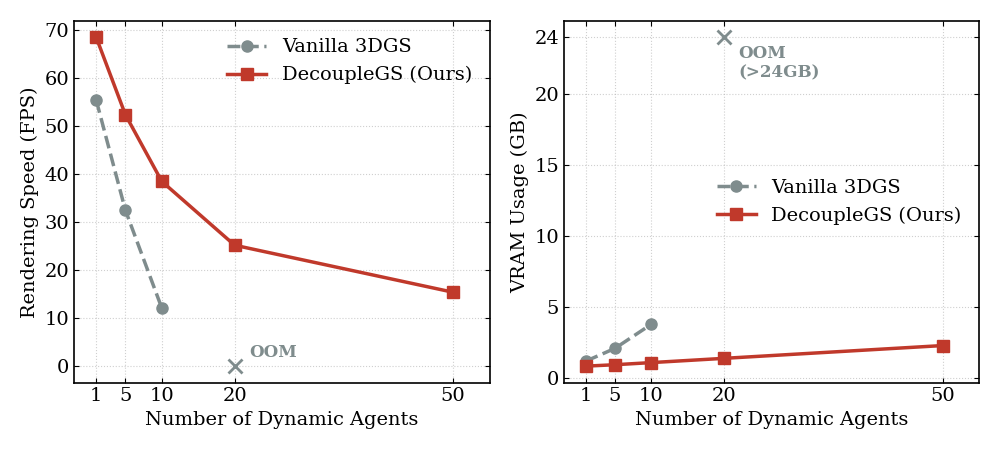}
  \caption{Scalability evaluation. DecoupleGS avoids the OOM bottlenecks of Vanilla 3DGS, demonstrating stable VRAM growth and usable frame rates up to 50 agents.}
  \label{fig:scalability}
\end{figure}

\textbf{Implementation and closed-loop protocol.}
All experiments are conducted on a single NVIDIA RTX 4090 GPU with 24 GB VRAM using custom CUDA kernels. The static background is optimized for 30k iterations, and the asset compression threshold $\tau_s$ is set to 0.005. All closed-loop evaluations are fully interactive. At each simulation step, DecoupleGS renders multi-view images for the tested planner, the E2E model predicts the ego trajectory, and background agents react online via coupled IDM and MOBIL behavioral models. The global scene state is then synchronized and re-rendered for the next step.

\begin{figure}[!t]
  \centering
  \includegraphics[width=\linewidth]{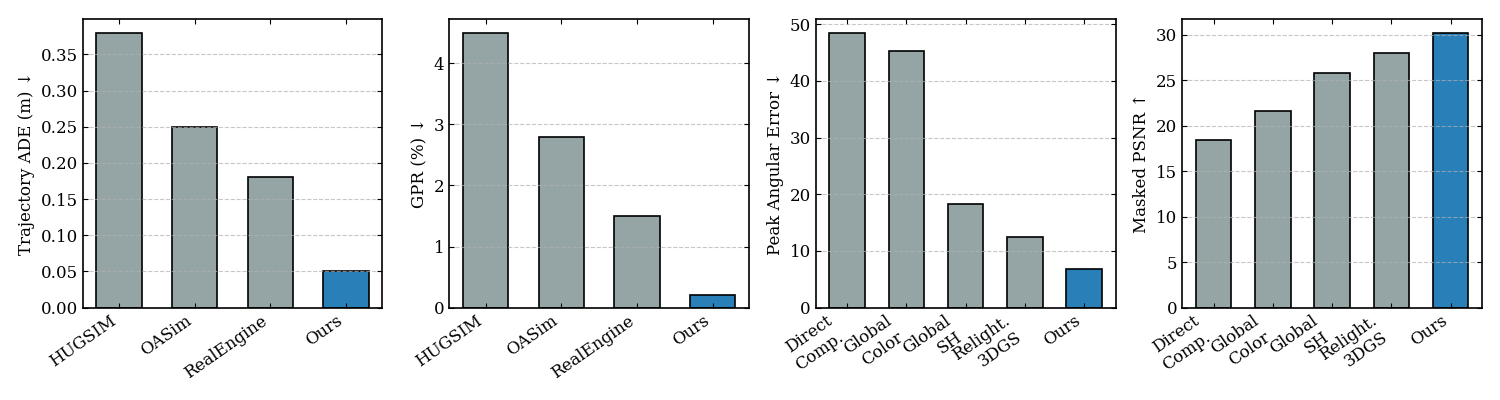}
  \caption{Quantitative comparisons of geometric registration and photometric realism against baseline methods. DecoupleGS achieves lower trajectory drift, fewer ground-penetration artifacts, and more consistent relighting.}
  \label{fig:geo_photo}
\end{figure}

\begin{figure}[!t]
  \centering
  \includegraphics[width=0.8\linewidth]{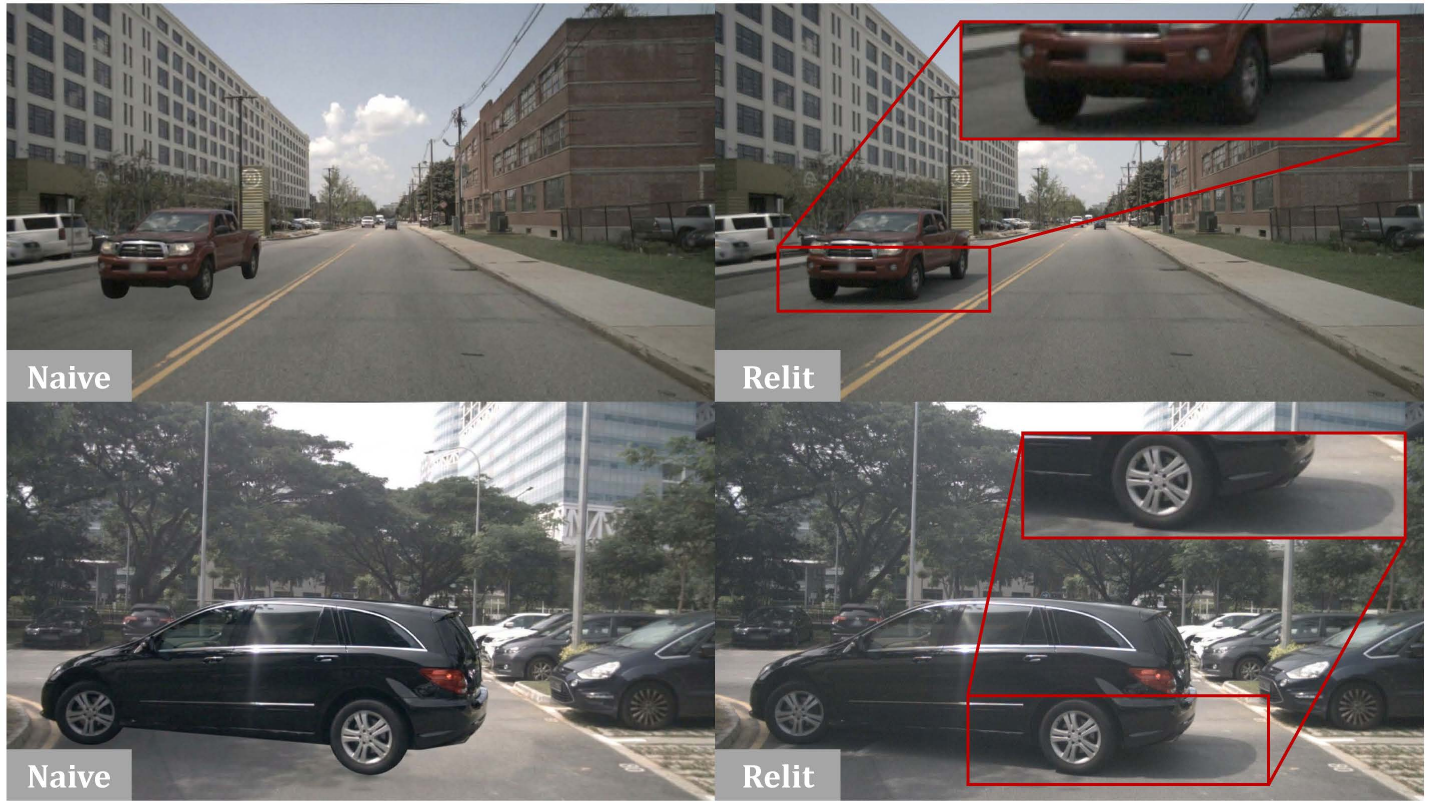}
  \caption{Qualitative comparison of the relighting module, demonstrating seamless integration of dynamic assets under various illuminations.}
  \label{fig:relighting}
\end{figure}

\subsection{Rendering Efficiency and Scalability}

\textbf{Asset compression.}
Table \ref{tab:compression} reports results under three traffic densities: Size S with 1--2 vehicles, Size M with 3--5 vehicles, and Size L with 6--10+ vehicles. Vanilla 3DGS preserves the highest rendering fidelity but becomes inefficient as traffic density increases, while LightGaussian achieves the most aggressive compression and fastest rendering at the cost of lower scene fidelity. DecoupleGS is designed for a different objective: it preserves the high-fidelity static background and compresses only dynamic assets, yielding the most balanced fidelity-efficiency trade-off. As shown in the final column, among the compared methods, DecoupleGS is the only one that consistently remains within the top-2 across all fidelity and efficiency metrics under every traffic density, which is more suitable for interactive E2E testing than optimizing a single isolated metric.

\textbf{Scalability.}
We further conduct a stress test by increasing the number of dynamic agents to 50. As shown in Fig. \ref{fig:scalability}, Vanilla 3DGS suffers rapid memory growth and encounters out-of-memory failures before reaching dense traffic. In contrast, DecoupleGS relies on a persistent background and compact VQ-indexed assets, leading to approximately linear memory growth. It maintains usable throughput even under extreme multi-agent settings, which is essential for large-scale closed-loop policy evaluation.

\subsection{Geometric and Photometric Fidelity}

\textbf{Map-guided geometric consistency.}
Fig. \ref{fig:geo_photo} evaluates the metric alignment of inserted agents. Compared with HUGSIM and RealEngine, our map-guided registration significantly reduces trajectory drift and ground penetration. By combining trajectory-to-lane sequence alignment, SE(2) planar correction, and opacity-accumulated vertical grounding, DecoupleGS achieves accurate lane-level placement while maintaining physically plausible contact with the road surface.

\textbf{Photometric realism.}
The same figure also reports photometric realism under held-out vehicle re-insertion. Our proxy-based relighting obtains the lowest PAE and competitive Masked PSNR by transferring local ambient illumination to canonical assets without online neural inference. As shown in Fig. \ref{fig:relighting}, inserted vehicles inherit scene-dependent lighting and contact shadows, avoiding the common mismatch between foreground appearance and background illumination.

\begin{table}[!t]
  \caption{Open-loop sim-to-real consistency and closed-loop E2E simulator comparison. 
  Open-loop results measure UniAD behavior under rendered observations, using real images as the reference input. 
  Bold values indicate the best result among simulated inputs. 
  Closed-loop scores are normalized to $[0,1]$ for consistency with Table \ref{tab:e2e}.}
  \label{tab:sim_consistency}
  \centering
  \resizebox{\linewidth}{!}{
  \begin{tabular}{@{}llccccc@{}}
    \toprule
    Setting & Metric & Real Input & Ours & HUGSIM~\cite{zhou2023hugsim} & RealEngine~\cite{jiang2025realengine} & OASim~\cite{yan2024oasim} \\
    \midrule
    \multirow{2}{*}{Open-loop}
    & mADE (m) $\downarrow$ & 0.76 & \textbf{0.82} & 0.94 & 1.15 & 1.02 \\
    & minTTC (s) $\uparrow$ & 3.5 & \textbf{3.3} & 2.8 & 2.4 & 2.9 \\
    \midrule
    \multirow{4}{*}{Closed-loop}
    & Driving Score $\uparrow$ & -- & \textbf{0.884} & 0.765 & 0.682 & 0.748 \\
    & Route Completion $\uparrow$ & -- & \textbf{0.956} & 0.814 & 0.795 & 0.851 \\
    & minTTC (s) $\uparrow$ & -- & \textbf{3.3} & 2.3 & 2.6 & 2.8 \\
    & Rendering FPS $\uparrow$ & -- & \textbf{45} & 12 & 32 & 18 \\
    \bottomrule
  \end{tabular}
  }
\end{table}

\begin{table}[!t]
  \caption{Closed-loop E2E AD testing across scenario difficulty levels. Values are Mean $\pm$ Std over 50 independent episodes.}
  \label{tab:e2e}
  \centering
  \small
  \setlength{\tabcolsep}{4.0pt}
  \renewcommand{\arraystretch}{0.95}
  \begin{tabular}{@{}llcccc@{}}
    \toprule
    Difficulty & Method & DS & SR & RC & minTTC \\
    \midrule
    Easy & UniAD~\cite{hu2023planning} & 0.725$\pm$0.04 & 0.850$\pm$0.03 & 0.780$\pm$0.05 & 0.812$\pm$0.06 \\
    & VAD~\cite{jiang2023vad} & 0.680$\pm$0.05 & 0.815$\pm$0.04 & 0.720$\pm$0.06 & 0.785$\pm$0.05 \\
    \midrule
    Medium & UniAD~\cite{hu2023planning} & 0.540$\pm$0.06 & 0.710$\pm$0.05 & 0.610$\pm$0.07 & 0.650$\pm$0.08 \\
    & VAD~\cite{jiang2023vad} & 0.450$\pm$0.07 & 0.630$\pm$0.06 & 0.520$\pm$0.08 & 0.580$\pm$0.07 \\
    \midrule
    Hard & UniAD~\cite{hu2023planning} & 0.380$\pm$0.08 & 0.520$\pm$0.07 & 0.450$\pm$0.09 & 0.480$\pm$0.10 \\
    & VAD~\cite{jiang2023vad} & 0.250$\pm$0.09 & 0.410$\pm$0.08 & 0.320$\pm$0.10 & 0.390$\pm$0.09 \\
    \midrule
    Extreme & UniAD~\cite{hu2023planning} & 0.195$\pm$0.07 & 0.310$\pm$0.09 & 0.250$\pm$0.08 & 0.280$\pm$0.08 \\
    & VAD~\cite{jiang2023vad} & 0.120$\pm$0.05 & 0.220$\pm$0.07 & 0.180$\pm$0.06 & 0.210$\pm$0.06 \\
    \bottomrule
  \end{tabular}
\end{table}

\subsection{Sim-to-real Consistency}

A reliable simulator should not only render visually plausible images but also preserve the behavior of downstream planners. We therefore conduct an open-loop sim-to-real consistency test using UniAD. Real images serve as the reference input, while different simulators provide rendered observations for the same driving scenes. As shown in Table \ref{tab:sim_consistency}, DecoupleGS most closely matches the real-data behavior in both mADE and minTTC, indicating a smaller sim-to-real behavioral gap than existing neural simulators.

\begin{figure}[!t]
  \centering
  \includegraphics[width=0.8\linewidth]{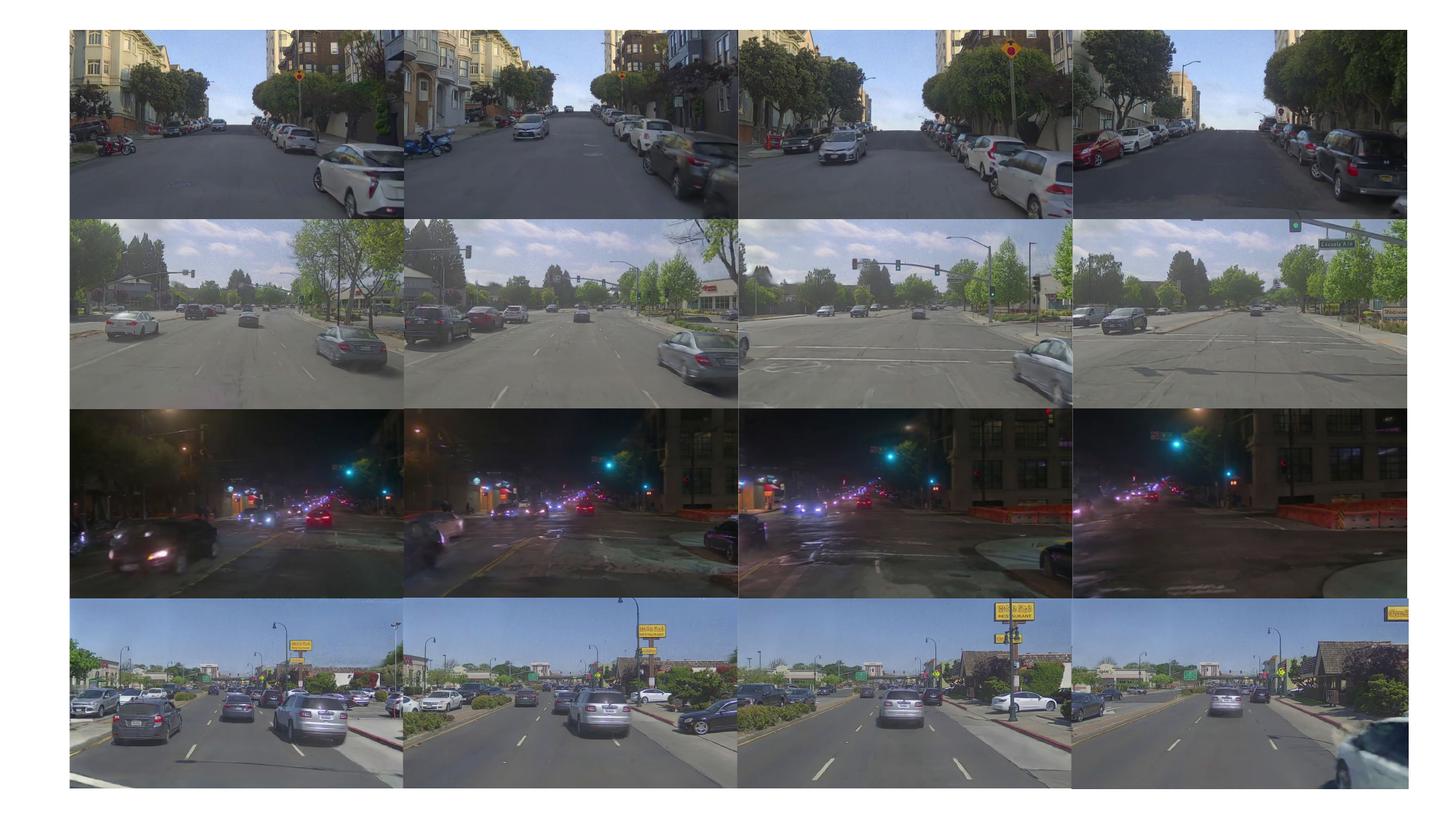}
  \caption{Representative camera views of UniAD evaluated across generated scenarios of increasing difficulty, from Easy to Extreme.}
  \label{fig:uniad}
\end{figure}

\begin{table}[!t]
  \caption{System-level ablation study evaluated on medium-density scenarios. 
  Values in parentheses indicate changes relative to the full framework.}
  \label{tab:ablation}
  \centering
  \small
  \setlength{\tabcolsep}{5.0pt}
  \renewcommand{\arraystretch}{1.03}
  \begin{tabular}{@{}lccc@{}}
    \toprule
    Methods & FPS $\uparrow$ & Trajectory ADE $\downarrow$ & Peak Angular Error $\downarrow$ \\
    \midrule
    w/o asset compression 
      & 8.5 {\scriptsize($\downarrow$26.9)} 
      & 0.05 {\scriptsize(=)} 
      & 6.8 {\scriptsize(=)} \\
    w/o map-guided reg. 
      & 36.2 {\scriptsize($\uparrow$0.8)} 
      & 0.48 {\scriptsize($\uparrow$0.43)} 
      & 6.8 {\scriptsize(=)} \\
    w/o relighting 
      & 38.5 {\scriptsize($\uparrow$3.1)} 
      & 0.05 {\scriptsize(=)} 
      & 48.5 {\scriptsize($\uparrow$41.7)} \\
    \rowcolor{oursrow}
    \textbf{Full Framework (Ours)}
      & \textbf{35.4}
      & \textbf{0.05}
      & \textbf{6.8} \\
    \bottomrule
  \end{tabular}
\end{table}

\subsection{Closed-loop E2E Testing}

\textbf{Simulator-level comparison.}
Beyond open-loop consistency, Table \ref{tab:sim_consistency} compares different simulators under interactive closed-loop testing. DecoupleGS achieves the highest normalized Driving Score and Route Completion while maintaining the largest safety margin measured by minTTC. It also runs at 45 FPS, substantially faster than HUGSIM and OASim. These results demonstrate that DecoupleGS is not merely a rendering module, but a practical closed-loop sensor simulation engine for E2E policy evaluation.

\textbf{Difficulty-based stress testing.}
We further evaluate UniAD and VAD across four scenario difficulty levels, as shown in Table \ref{tab:e2e}, with representative closed-loop visualizations provided in Fig. \ref{fig:uniad}. The difficulty is controlled by traffic density, illumination, and background-agent aggressiveness. Easy scenarios contain sparse traffic and clear daytime illumination. Medium scenarios introduce moderate density and standard urban maneuvers. Hard scenarios include dense traffic and challenging illumination such as dusk or heavy overcast. Extreme scenarios further introduce adversarial lighting and safety-critical behaviors such as sudden braking and aggressive cut-ins.

Both planners degrade as the scenario difficulty increases. UniAD maintains a relatively high completion rate in Easy scenarios but suffers under severe occlusion and dense interactions, while VAD is more sensitive to dense multi-agent constraints, causing a steeper drop in Driving Score under Hard and Extreme settings. The qualitative examples in Fig. \ref{fig:uniad} further illustrate that DecoupleGS can synthesize visually diverse and increasingly challenging closed-loop scenarios, including dense traffic, low-light conditions, and safety-critical interactions. These results show that DecoupleGS can generate structured and controllable stress tests that expose planner vulnerabilities.

\begin{figure}[!t]
  \centering
  \includegraphics[width=\linewidth]{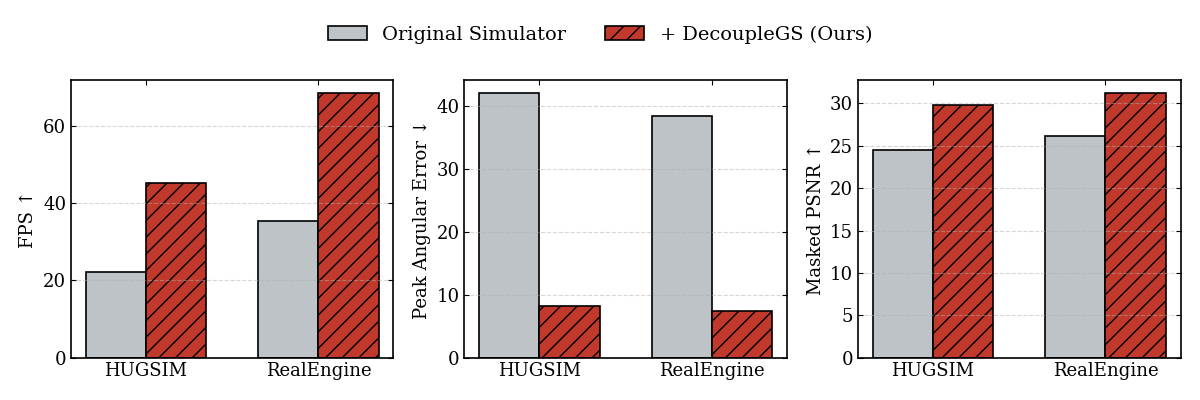}
  \caption{Plug-and-play capability. Integrating our modules into existing simulators improves rendering speed and photometric accuracy.}
  \label{fig:plug_and_play}
\end{figure}

\subsection{Ablation Studies}

Table \ref{tab:ablation} decomposes the effect of each core module under medium-density scenarios. Without asset compression, throughput collapses from 35.4 FPS to 8.5 FPS, making interactive multi-agent simulation impractical. Without map-guided registration, ADE increases from 0.05 m to 0.48 m, showing that neural coordinates alone do not guarantee metric-consistent placement. Without relighting, PAE increases from $6.8^{\circ}$ to $48.5^{\circ}$, exposing severe foreground-background illumination mismatch. The full framework therefore offers the best balanced trade-off across efficiency, geometric alignment, and photometric consistency.

\textbf{Plug-and-play capability.}
We further validate that the proposed modules are solver-agnostic. As shown in Fig. \ref{fig:plug_and_play}, integrating DecoupleGS components into existing simulators improves runtime and photometric consistency, indicating that our compression, registration, and relighting modules can serve as general-purpose building blocks for interactive neural driving simulation.

\subsection{Limitations} While DecoupleGS enables efficient and consistent closed-loop testing, it still has limitations. First, its proxy-based relighting favors runtime efficiency over physically exact global illumination, and may struggle with complex multi-bounce lighting, specular effects, and cast shadows. Second, map-guided registration relies on reliable semantic topology, making multi-level roads, overpasses, under-bridge regions, and poorly detected lanes more challenging. Third, the current framework mainly targets rigid traffic agents and camera simulation. Future work will explore non-rigid foregrounds, full background compression, richer interactive behaviors, and unified camera--LiDAR sensor simulation.

\section{Conclusion} \label{sec:conclusion} This paper presents DecoupleGS, a decoupled 3DGS framework for interactive E2E autonomous driving testing. By separating persistent backgrounds from manipulable agents, DecoupleGS addresses efficiency, geometric, and photometric conflicts through asset compression, map-guided registration, and proxy-based relighting. Experiments demonstrate a balanced fidelity-efficiency trade-off, improved sim-to-real consistency, and fully interactive closed-loop policy evaluation. Future work will extend DecoupleGS toward richer illumination, non-rigid agents, and unified RGB-LiDAR simulation.

\section*{Acknowledgements}
This work was supported by the National Natural Science Foundation of China (Grant No. 52502514) and the Fundamental Research Funds for the Central Universities.

\bibliographystyle{splncs04}
\bibliography{references_arxiv}

\clearpage

\markboth{Supplementary Material}{Supplementary Material}

\begin{center}
  {\Large\bfseries
  DecoupleGS: Interactive 3D Gaussian Splatting for
  End-to-End Autonomous Driving Testing\par}
  \vspace{0.65em}
  {\large\bfseries Supplementary Material\par}
\end{center}
\vspace{1em}

\section*{Overview}

This supplementary document provides additional details, theoretical expansions, and extended experimental results to support the main paper \textit{DecoupleGS: Interactive 3D Gaussian Splatting for End-to-End Autonomous Driving Testing}. The material is structured as follows:

\begin{itemize}
    \item \textbf{Section \ref{sec:implementation}}: Detailed implementation configurations, including system overhead, training schedules, specific hyperparameters for both background reconstruction and asset compression, and dataset curation strategies.
    \item \textbf{Section \ref{sec:method_expansions}}: In-depth mathematical formulations for the Vector Quantization (VQ) optimization, proxy-based relighting calibration, and contact shadow generation, alongside comprehensive algorithmic pseudocodes for the offline and online pipelines.
    \item \textbf{Section \ref{sec:experimental_setups}}: Concrete definitions of the evaluation scenario difficulties (Easy to Extreme), rigorous mathematical formulations of the evaluation metrics, and complete details regarding the interactive closed-loop system integration, including the coupled Intelligent Driver Model (IDM) \cite{treiber2000congested} and Minimizing Overall Braking Induced by Lane change (MOBIL) \cite{kesting2007general} behavioral models for background agents.
    \item \textbf{Section \ref{sec:additional_results}}: Extended qualitative results demonstrating the framework's robust capabilities in multi-view scene decoupling and interactive scene editing.
\end{itemize}

\appendix


\section{Implementation Details}
\label{sec:implementation}

In this section, we provide the comprehensive engineering configurations required to reproduce the DecoupleGS framework, elaborating on the concise descriptions presented in Section 4.2 of the main text.

\begin{table}[t]
  \caption{Hyperparameters for Static Background 3DGS Optimization.}
  \label{tab:hyper_bg}
  \centering
  \begin{tabular}{@{}ll@{}}
    \toprule
    \textbf{Parameter} & \textbf{Value} \\
    \midrule
    Total Iterations & 30,000 \\
    Position Learning Rate ($\mu$) & $1.6 \times 10^{-4}$ decaying to $1.6 \times 10^{-6}$ \\
    Feature Learning Rate ($c$) & $2.5 \times 10^{-3}$ \\
    Opacity Learning Rate ($\alpha$) & $0.05$ \\
    Scaling Learning Rate ($s$) & $5.0 \times 10^{-3}$ \\
    Densification Interval & Every 100 iterations (from iter 500 to 15,000) \\
    Densification Gradient Threshold & $2.0 \times 10^{-4}$ \\
    Opacity Reset Interval & Every 3,000 iterations \\
    \bottomrule
  \end{tabular}
\end{table}

\subsection{Training Schedules and System Overhead}

Our pipeline consists of two primary offline optimization stages: background scene reconstruction and canonical asset compression. The framework is implemented in PyTorch, utilizing custom CUDA kernels for efficient unified rasterization and bounding-box level frustum culling. All optimizations and evaluations are conducted on a single NVIDIA RTX 4090 GPU (24 GB VRAM) alongside an Intel Core i9-13900K CPU.

\noindent\textbf{Static Background Reconstruction.}
The background environment is optimized using the vanilla 3D Gaussian Splatting (3DGS) \cite{kerbl20233d} pipeline for 30,000 iterations. We employ the Adam optimizer with a dynamic learning rate. Table \ref{tab:hyper_bg} summarizes the detailed hyperparameter settings. On average, optimizing a single background scenario (approx. 200 frames) takes roughly 45 minutes.

\noindent\textbf{Semantic-aware Asset Compression.}
For dynamic traffic agents, canonical Gaussians are first optimized for 20,000 iterations. Unlike the unbounded background, vehicle primitives are strictly constrained within a scaled 3D bounding box to prevent floaters. The densification gradient threshold is slightly relaxed to $4.0 \times 10^{-4}$ to avoid excessive primitive generation inside the vehicle cabin. Offline training per vehicle asset takes approximately 15 minutes.

Subsequently, explicit importance scoring and adaptive pruning are applied. The pruning threshold $\tau_{\text{s}}$ is strictly set to 0.005. The weighting coefficients in Equation (4) of the main text were determined via grid search over the discrete space $\{0.1, 0.3, 0.5, 0.7, 0.9\}$. The optimal configuration was found to be: $w_{\text{vis}} = 0.5$, $w_{\text{col}} = 0.3$, and $w_{\text{ent}} = 0.2$. This configuration strongly prioritizes the preservation of visible structural contours while moderately suppressing internal unobservable primitives.

For attribute VQ, we maintain codebooks of size $K_{\text{color}} = 1024$ for Spherical Harmonics (SH) features and $K_{\text{shape}} = 512$ for covariances. Crucially, to ensure convergence, the codebooks are initialized using K-Means clustering over the pruned primitives rather than random uniform sampling. They are then dynamically updated during the final 5,000 iterations using Exponential Moving Average (EMA) with a momentum factor of $\gamma = 0.99$.

\subsection{Map-guided Registration Parameters}
For the geometric registration module (Section 3.3 in the main text), the Dynamic Time Warping (DTW) cost matrix $C_{t,u}$ is constructed by balancing Euclidean distance and heading mismatch. The scalar weight for the heading penalty is empirically set to $\lambda_{\text{heading}} = 2.5$, ensuring that vehicles strictly adhere to lane orientations during sharp turns or lane changes. For scenarios where HD-Map are not provided, MapTRv2 \cite{liao2025maptrv2} is deployed with its default pre-trained weights on nuScenes \cite{caesar2020nuscenes}, and the extracted polylines are rasterized at a 0.1m/pixel resolution for the Orthogonal Procrustes analysis.

\subsection{Dataset Curations}

To rigorously evaluate DecoupleGS across diverse and challenging conditions, we carefully curated subsets from multiple large-scale autonomous driving datasets.

\noindent\textbf{Background Scenarios.} 
We selected 15 dynamic clips (each roughly 20 seconds at 10 Hz) from the \textbf{nuScenes} dataset, explicitly choosing high-complexity sequences. Additionally, we extracted 10 sequences from \textbf{PandaSet} \cite{xiao2021pandaset} to cover adversarial illumination conditions (dusk, night, and overcast). To obtain clean background plates, dynamic objects within these sequences were masked out using the Segment Anything Model (SAM) \cite{kirillov2023segment} with a morphological dilation of 5 pixels to prevent edge artifacts.

\noindent\textbf{Manipulable Canonical Assets.} 
The canonical vehicle instances were sourced from the \textbf{3DRealCar} \cite{du20253drealcar} dataset. We systematically sampled 20 distinct vehicle models to ensure a diverse morphological distribution: Sedans (10 instances), SUVs (6 instances), and Large Commercial Vehicles (4 instances).

\section{Methodological Expansions}
\label{sec:method_expansions}

In this section, we provide the complete mathematical formulations for the VQ optimization and the proxy-based relighting operators, which were briefly introduced in Section 3 of the main manuscript.

\subsection{Vector Quantization Optimization via EMA}
\label{subsec:vq_ema}

As described in Section 3.2 of the main text, to overcome the memory bottleneck of storing multiple vehicle instances, we apply VQ to the high-dimensional attributes (covariance $\Sigma_i$ and SH color coefficients $c_i$). Instead of a static quantization, our codebooks $\mathcal{C}_{\text{shape}}$ and $\mathcal{C}_{\text{color}}$ are dynamically optimized using an EMA to prevent index collapse and maintain representation fidelity.

Let $e_k^{(t)}$ denote the $k$-th entry in the codebook at iteration $t$, and $x_i$ denote the input continuous attribute of the $i$-th Gaussian. In the forward pass, each primitive is assigned to its nearest codebook entry:
\begin{equation}
    q_i = \arg\min_k || x_i - e_k^{(t)} ||_2^2
\end{equation}
To update the codebook without requiring differentiable backpropagation through the argmin operation, we track the cluster size $N_k$ and the sum of assigned vectors $m_k$. At each optimization step, we update these statistics using EMA with a momentum factor $\gamma \in [0, 1)$ (we set $\gamma = 0.99$):
\begin{align}
    N_k^{(t)} &= \gamma N_k^{(t-1)} + (1 - \gamma) \sum_i \mathbb{I}(q_i = k) \\
    m_k^{(t)} &= \gamma m_k^{(t-1)} + (1 - \gamma) \sum_{i : q_i = k} x_i
\end{align}
where $\mathbb{I}(\cdot)$ is the indicator function. The codebook entry is then explicitly updated as the smoothed centroid:
\begin{equation}
    e_k^{(t)} = \frac{m_k^{(t)}}{N_k^{(t)}}
\end{equation}
To handle dead codes (entries where $N_k^{(t)} \approx 0$), we periodically re-initialize them using K-Means++ over a random subset of the current active primitives $x_i$, ensuring maximum utilization of the codebook capacity.

\subsection{Proxy-based Relighting Analytical Calibration}
\label{subsec:relighting_math}

A core contribution of DecoupleGS is the ability to transfer ambient illumination to canonical assets without online MLP inference. In Equation (8) of the main text, we defined the linear modulation:
\begin{equation}
    c_{\text{out}} = \mathcal{T}(\mathcal{L}(x)) \otimes c_{\text{can}} + b(\mathcal{L}(x))
\end{equation}
Here, we clarify the derivation of the operators $\mathcal{T}(\cdot)$ and $b(\cdot)$. To maintain real-time performance, we parameterize these operators as affine functions of the local ambient descriptor $\mathcal{L}(x) \in \mathbb{R}^{27}$ (aggregating the first 3 bands of SH coefficients from the background):
\begin{equation}
    \mathcal{T}(\mathcal{L}) = \mathbf{W}_T \mathcal{L} + \mathbf{w}_0, \quad b(\mathcal{L}) = \mathbf{W}_B \mathcal{L} + \mathbf{b}_0
\end{equation}
To avoid online optimization, we perform an offline analytical calibration. We generate a synthetic calibration dataset by rendering the canonical asset under $N$ diverse, pre-captured HDRI lighting probes. For each primitive $i$, we extract its canonical color $c_{\text{can}, i}$ and its ground-truth relit color $c_{\text{out}, i}^{(n)}$ under the $n$-th probe descriptor $\mathcal{L}^{(n)}$. 

We then formulate this as a global Ordinary Least Squares (OLS) regression problem over the dataset $\{ c_{\text{can}, i}, \mathcal{L}^{(n)}, c_{\text{out}, i}^{(n)} \}$. The objective is to find the optimal global parameter matrices $\Theta = \{\mathbf{W}_T, \mathbf{w}_0, \mathbf{W}_B, \mathbf{b}_0\}$ that minimize the photometric residual:
\begin{equation}
    \min_{\Theta} \sum_{n=1}^N \sum_{i \in \mathcal{V}} \left|\left| \left( \mathbf{W}_T \mathcal{L}^{(n)} + \mathbf{w}_0 \right) \otimes c_{\text{can}, i} + \left( \mathbf{W}_B \mathcal{L}^{(n)} + \mathbf{b}_0 \right) - c_{\text{out}, i}^{(n)} \right|\right|_2^2
\end{equation}
Since this objective is quadratic with respect to the parameters in $\Theta$, it admits a closed-form analytical solution via the normal equations. Once computed offline, $\Theta$ is stored globally. During online E2E testing, relighting a vehicle only requires two efficient matrix-vector multiplications per vehicle frame, completely bypassing the computational bottleneck of evaluating neural networks on millions of primitives.

\subsection{Parametric Contact Shadow Generation}

To ground the vehicle visually, Equation (9) introduces the shadow modulation function $1 - \lambda \cdot I_{\text{dom}} \cdot M_{\text{shadow}}(u)$. We define the spatial shadow mask $M_{\text{shadow}}(u)$ as a 2D super-ellipse projected onto the estimated ground plane, smoothly decaying outward. Let $u = (u_x, u_y)$ be the local ground-plane coordinates relative to the vehicle center, and $(L, W)$ be the length and width of the vehicle's bounding box footprint. The mask is defined as:
\begin{equation}
    M_{\text{shadow}}(u) = \exp \left( - \left( \left| \frac{u_x}{L/2} \right|^p + \left| \frac{u_y}{W/2} \right|^p \right)^{\frac{1}{p}} \cdot \tau_{\text{decay}} \right)
\end{equation}
where $p = 4$ creates a rounded rectangular shape matching the vehicle chassis, and $\tau_{\text{decay}}$ controls the softness of the shadow edge. The dominant light intensity $I_{\text{dom}} = \max(0, \mathcal{L}_{\text{DC}}(x))$ dynamically scales this mask, ensuring shadows are stark under direct sunlight (high $I_{\text{dom}}$) and faint in overcast or night scenarios (low $I_{\text{dom}}$).

\begin{algorithm}[h]
\caption{Offline Semantic-aware Asset Compression}
\label{alg:offline}
\begin{algorithmic}[1]
\REQUIRE Raw canonical 3DGS vehicle $\mathcal{V}_{\text{raw}}$, pruning threshold $\tau_{\text{s}}$, EMA momentum $\gamma$
\ENSURE Compact canonical volume $\mathcal{V}_{\text{compact}}$ and Codebooks $\mathcal{C}_{\text{shape}}, \mathcal{C}_{\text{color}}$
\STATE Initialize codebooks $\mathcal{C}_{\text{shape}}, \mathcal{C}_{\text{color}}$ via K-Means++ on raw attributes
\FOR{each primitive $g_i \in \mathcal{V}_{\text{raw}}$}
    \STATE Calculate visibility $V_i$, color contrast $D_i$, and texture entropy $H_i$
    \STATE Compute scalar importance $s_i = w_{\text{vis}}V_i + w_{\text{col}}D_i + w_{\text{ent}}H_i$
    \IF{$s_i < \tau_{\text{s}}$}
        \STATE Prune $g_i$ from $\mathcal{V}_{\text{raw}}$
    \ENDIF
\ENDFOR
\FOR{EMA iteration $t = 1$ to $5000$}
    \FOR{each retained primitive $g_i$}
        \STATE Find nearest indices $q_i^{\text{shape}}, q_i^{\text{color}}$ in $\mathcal{C}_{\text{shape}}, \mathcal{C}_{\text{color}}$
        \STATE Update cluster assignments and sizes
    \ENDFOR
    \STATE Update $\mathcal{C}_{\text{shape}}, \mathcal{C}_{\text{color}}$ using EMA with momentum $\gamma$
\ENDFOR
\STATE Store $\mathcal{V}_{\text{compact}}$ using quantized indices $q_i^{\text{shape}}, q_i^{\text{color}}$
\end{algorithmic}
\end{algorithm}

\subsection{Algorithmic Overview of the DecoupleGS Pipeline}
\label{subsec:pseudocode}

To provide a clear, step-by-step understanding of our decoupled framework, we summarize the core operations in two algorithms. Algorithm \ref{alg:offline} details the offline semantic-aware asset compression. Algorithm \ref{alg:online} outlines the online closed-loop scenario generation, demonstrating how map-guided registration and proxy-based relighting seamlessly integrate dynamic agents into the unified rasterizer without neural inference.

\begin{algorithm}[h]
\caption{Online Decoupled Integration \& Unified Rendering}
\label{alg:online}
\begin{algorithmic}[1]
\REQUIRE Static background $\mathcal{S}_{\text{bg}}$, HD Map Lane $\mathcal{L}$, Asset Library $\{\mathcal{V}_k\}$, Planned Trajectory $\mathcal{T}=\{p_t\}$
\ENSURE Synthesized image $I_t$ for E2E policy evaluation
\STATE \textbf{Map-guided Registration:}
\STATE Compute alignment cost matrix $C_{t,u}$ between $\mathcal{T}$ and $\mathcal{L}$
\STATE Extract optimal warping path via DTW and compute 2D rigid transform $T_{\text{align}} \in SE(2)$
\STATE Calculate opacity-weighted vertical height $z_{\text{asset}}$ via ray accumulation
\STATE Output fully registered 6-DoF pose $\mathcal{T}_k(t) = [\mathbf{R}_k | \mathbf{t}_k]$
\STATE \textbf{Proxy-based Relighting:}
\STATE Sample local ambient descriptor $\mathcal{L}(x)$ from neighboring $\mathcal{S}_{\text{bg}}$ primitives
\STATE Modulate asset SH coefficients: $c_{\text{out}} \leftarrow \mathcal{T}(\mathcal{L}(x)) \otimes c_{\text{can}} + b(\mathcal{L}(x))$
\STATE Generate dynamic shadow mask $M_{\text{shadow}}$ scaled by $I_{\text{dom}}$
\STATE \textbf{Unified Rasterization:}
\STATE Apply spatial transforms: $\mu^{(W)} \leftarrow \mathbf{R}_k \mu^{(L)} + \mathbf{t}_k$, $\Sigma^{(W)} \leftarrow \mathbf{R}_k \Sigma^{(L)} \mathbf{R}_k^\top$
\STATE Aggregate all primitives: $\mathcal{P} \leftarrow \mathcal{S}_{\text{bg}} \cup \mathcal{V}_k$
\STATE Perform depth-sorting and continuous $\alpha$-blending to output frame $I_t$
\end{algorithmic}
\end{algorithm}

\section{Extended Experimental Setups}
\label{sec:experimental_setups}

In Section 4.5 of the main text, we presented the closed-loop evaluation of two state-of-the-art End-to-End (E2E) autonomous driving planners, UniAD \cite{hu2023planning} and VAD \cite{jiang2023vad}. To fully demonstrate the robustness and interactivity of our DecoupleGS framework, we detail the scenario definitions, the mathematical formulation of our metrics, the background agent behavioral engine, and the complete interactive closed-loop evaluation pipeline in this section.



\subsection{Definition of Evaluation Scenarios}
\label{subsec:scenario_definitions}

To systematically stress-test the E2E models, we categorized the generated simulation scenarios into four distinct difficulty levels based on traffic density, environmental illumination, and behavioral complexity of the background agents. Each episode runs for up to 20 seconds at a control frequency of 2Hz (matching the standard nuScenes planner configuration).

\begin{itemize}
    \item \textbf{Easy:} Clear daytime illumination with favorable visibility. Sparse traffic density (1 to 3 dynamic agents within a 50m radius). The background agents exhibit stable, predictable behaviors (e.g., constant speed lane keeping).
    \item \textbf{Medium:} Daytime or overcast illumination causing mild shadow artifacts. Moderate traffic density (4 to 8 dynamic agents). Scenarios include standard urban maneuvers such as yielding at intersections and smooth lane changes by background traffic.
    \item \textbf{Hard:} Challenging illumination conditions, including dusk, dawn, or heavy overcast skies, demanding robust perception. Dense traffic (9 to 15 dynamic agents). Background agents perform complex maneuvers, including congested intersection crossings and close-proximity cut-ins.
    \item \textbf{Extreme:} Adversarial lighting conditions such as night-time rendering with extreme low-light constraints or blinding specular glare. Extremely dense traffic ($>15$ dynamic agents). The background behavioral engine deliberately injects safety-critical behaviors, such as sudden hard braking by the lead vehicle or aggressive swerving, forcing the ego-vehicle into high-risk interactions.
\end{itemize}

\subsection{Detailed Evaluation Metrics Formulation}
\label{subsec:metrics_detail}

As absolute 3D ground truth is intractable for in-the-wild datasets, we present the exact mathematical formulations of our targeted metrics to ensure rigorous evaluation.

\subsubsection{Photometric Realism Metrics.}
We employ a hold-out strategy using precise SAM masks $M$. Let $I(u)$ be the synthesized color at pixel $u$, and $\hat{I}(u)$ be the ground-truth color. 
\textbf{Masked Peak Signal-to-Noise Ratio (PSNR)} evaluates the fidelity strictly within the dynamic asset's footprint:
\begin{equation}
    \text{Masked PSNR} = 10 \cdot \log_{10} \left( \frac{\mathrm{MAX}_I^2}{\frac{1}{|M|} \sum_{u \in M} ||I(u) - \hat{I}(u)||_2^2} \right)
\end{equation}
\noindent\textbf{Peak Intensity Error (PIE)} measures the maximum absolute error in luminance (grayscale intensity) between the synthesized image and the ground-truth image. This captures the worst-case localized illumination magnitude mismatch. Let $Y(\cdot)$ denote the grayscale luminance channel extraction:
\begin{equation}
    \text{PIE} = \max_{u \in M} \left| Y(I(u)) - Y(\hat{I}(u)) \right|
\end{equation}
\textbf{Peak Angular Error (PAE)} explicitly isolates environmental illumination color shifts from pure intensity changes:
\begin{equation}
    \text{PAE} = \max_{u \in M} \arccos \left( \frac{I(u) \cdot \hat{I}(u)}{||I(u)||_2 ||\hat{I}(u)||_2} \right)
\end{equation}

\subsubsection{Geometric Alignment Metrics.}
\textbf{Trajectory Average Displacement Error (ADE)} measures the lateral drift. Given the 2D projected trajectory $\mathcal{T} = \{p_t\}_{t=1}^T$ and the semantic lane centerlines $\mathcal{L}$:
\begin{equation}
    \text{ADE} = \frac{1}{T} \sum_{t=1}^T \min_{l \in \mathcal{L}} ||p_t - l||_2
\end{equation}
\textbf{Ground Penetration Rate (GPR)} evaluates terrain alignment. For a vehicle with $N_a$ bottom anchor points, let $z_{j,t}$ be the predicted vertical height, and $z_{\text{ground}}(x,y)$ be the local opacity-accumulated ground plane:
\begin{equation}
    \text{GPR} = \frac{1}{T \cdot N_a} \sum_{t=1}^T \sum_{j=1}^{N_a} \mathbb{I} \left( z_{j,t} < z_{\text{ground}}(x_{j,t}, y_{j,t}) \right)
\end{equation}

\begin{algorithm}[t!]
\caption{IDM Acceleration Update}
\label{alg:idm}
\begin{algorithmic}[1]
\REQUIRE Ego state ($v, x, L_{\text{car}}$), Leader state ($v_{\text{lead}}, x_{\text{lead}}$), Parameters ($v_0, T, a_{\text{max}}, b_{\text{comf}}, \delta, s_0$)
\ENSURE Longitudinal acceleration $a_{\text{out}}$
\IF{Leader exists in the current lane}
    \STATE $s \leftarrow \max(x_{\text{lead}} - x - L_{\text{car}}, 0.1)$ \COMMENT{Prevent zero-division in crash edge-cases}
    \STATE $\Delta v \leftarrow v - v_{\text{lead}}$
\ELSE
    \STATE $s \leftarrow 10000.0, \quad \Delta v \leftarrow 0.0$ \COMMENT{Free road assumption}
\ENDIF
\STATE $s^* \leftarrow s_0 + v T + \frac{v \Delta v}{2\sqrt{a_{\text{max}} b_{\text{comf}}}}$
\STATE $a_{\text{out}} \leftarrow a_{\text{max}} \left( 1 - \left(\frac{v}{v_0}\right)^\delta - \left(\frac{s^*}{s}\right)^2 \right)$
\RETURN $a_{\text{out}}$
\end{algorithmic}
\end{algorithm}

\subsubsection{E2E Policy Evaluation Metrics.}
\textbf{Route Completion (RC)} is the distance navigated $D_{\text{traveled}}$ divided by the total planned distance $D_{\text{planned}}$: $\text{RC} = D_{\text{traveled}} / D_{\text{planned}}$. 
\textbf{Driving Score (DS)} weights $\text{RC}$ by severe safety infractions (where $p_i \in (0, 1)$ is the penalty for infraction $i$ occurring $C_i$ times):
\begin{equation}
    \text{DS} = \text{RC} \times \prod_{i} p_i^{C_i}
\end{equation}
\textbf{Minimum Time-to-Collision (minTTC)} captures the strict safety margin under adversarial scenarios:
\begin{equation}
    \text{minTTC} = \min_{t \in [0, T]} \left( \frac{||x_{\text{ego}}(t) - x_{\text{agent}}(t)||_2}{v_{\text{rel}}(t)} \right)
\end{equation}

\begin{algorithm}[t!]
\caption{MOBIL Lane-Changing Decision}
\label{alg:mobil}
\begin{algorithmic}[1]
\REQUIRE Ego vehicle $C$, Current lane $L_{\text{curr}}$, Neighboring lanes $\{L_{\text{target}}\}$
\ENSURE Best lane choice $L_{\text{best}}$
\STATE $L_{\text{best}} \leftarrow L_{\text{curr}}$, $G_{\text{max}} \leftarrow -999.0$
\STATE Calculate current acceleration $a_c \leftarrow \mathrm{IDM}(C, \text{Leader}_{\text{curr}})$
\FOR{each feasible $L_{\text{target}}$}
    \STATE Identify New Leader ($\tilde{L}$) and New Follower ($\tilde{F}$) in $L_{\text{target}}$
    \STATE Calculate $\tilde{a}_{\text{n}} \leftarrow \mathrm{IDM}(\tilde{F}, C)$
    \IF{$\tilde{a}_{\text{n}} < -b_{\text{safe}}$}
        \STATE \textbf{continue} \COMMENT{Safety criterion violated, abort lane change}
    \ENDIF
    \STATE Calculate predicted ego acceleration $\tilde{a}_{\text{c}} \leftarrow \mathrm{IDM}(C, \tilde{L})$
    \STATE Calculate gain $G \leftarrow \tilde{a}_{\text{c}} - a_c$ \COMMENT{Simplified self-interest gain formulation}
    \IF{$G > a_{\text{thr}}$ \AND $G > G_{\text{max}}$}
        \STATE $G_{\text{max}} \leftarrow G$, $L_{\text{best}} \leftarrow L_{\text{target}}$
    \ENDIF
\ENDFOR
\RETURN $L_{\text{best}}$
\end{algorithmic}
\end{algorithm}

\subsection{Background Agent Behavioral Modeling}
\label{subsec:agent_modeling}

Unlike traditional simulators that rely on deterministic open-loop log replay, evaluating E2E policies requires background agents that exhibit realistic, reactive, and physically plausible behaviors. We empower the background traffic with a coupled longitudinal-lateral behavioral model driven by the IDM and the MOBIL model.

\subsubsection{Longitudinal Control (IDM).}
Background agents update their longitudinal acceleration $a_{\text{IDM}}$ based on current velocity $v$, leader velocity $v_{\text{lead}}$, and the gap $s$:
\begin{equation}
    a_{\text{IDM}}(s, v, \Delta v) = a_{\text{max}} \left[ 1 - \left( \frac{v}{v_0} \right)^\delta - \left( \frac{s^*(v, \Delta v)}{s} \right)^2 \right]
\end{equation}
where $\Delta v = v - v_{\text{lead}}$, $v_0$ is desired velocity, and $a_{\text{max}}$ is maximum acceleration. The desired minimum gap $s^*$ is computed as $s^* = s_0 + v T + \frac{v \Delta v}{2\sqrt{a_{\text{max}} b_{\text{comf}}}}$, detailed in Algorithm \ref{alg:idm}.

\subsubsection{Lateral Control (MOBIL).}
A lane change is executed if it satisfies a safety criterion (braking induced on the new follower $\tilde{a}_{\text{n}} \ge -b_{\text{safe}}$) and an incentive criterion:
\begin{equation}
    \underbrace{(\tilde{a}_{\text{c}} - a_c)}_{\text{ego gain}} + p \underbrace{\left( (\tilde{a}_{\text{n}} - a_n) + (\tilde{a}_{\text{o}} - a_o) \right)}_{\text{neighbors' gain}} > a_{\text{thr}}
\end{equation}
Algorithm \ref{alg:mobil} outlines this decision logic for complex urban maneuvers.

\subsection{Interactive Closed-loop System Integration}
\label{subsec:closed_loop_integration}

The core value of DecoupleGS lies in bridging photorealistic rendering with fully interactive simulation. Unlike open-loop testing, our framework creates a dynamic game-theoretic environment where the ego-vehicle (driven by the E2E policy) and background agents (driven by IDM-MOBIL) react to one another continuously. 

The closed-loop evaluation follows a synchronous step-by-step protocol. At each timestamp $t$, the simulator renders the multi-view sensor data based on the ego-vehicle's current 6-DoF pose. The E2E planner consumes these images to infer the surrounding environment and outputs a local planned trajectory for the ego-vehicle. Concurrently, the background agent engine perceives the ego-vehicle's current state and calculates the optimal longitudinal and lateral responses for all traffic participants. Finally, a global kinematic update resolves all trajectories to prevent physical overlap, advancing the entire scene to $t+1$. This fully interactive data flow is summarized in Algorithm \ref{alg:closed_loop}.

\begin{algorithm}[t!]
\caption{Interactive Closed-Loop E2E Evaluation Pipeline}
\label{alg:closed_loop}
\begin{algorithmic}[1]
\REQUIRE E2E Policy $\pi_{\text{ego}}$, Background Engine $\pi_{\text{bg}}$ (IDM+MOBIL), Static Background $\mathcal{S}_{\text{bg}}$, Dynamic Assets $\{\mathcal{V}_k\}$, Simulation Time $T_{\text{max}}$, Step $\Delta t$
\ENSURE Episode Driving Score ($\text{DS}$) and Minimum Time-to-Collision ($\text{minTTC}$)
\STATE Initialize $t \leftarrow 0$, Ego state $S_{\text{ego}}^{(0)}$, Background states $\mathbb{S}_{\text{bg}}^{(0)}$
\WHILE{$t < T_{\text{max}}$ \AND No Terminal Collision}
    \STATE \textbf{// 1. High-Fidelity Sensor Simulation}
    \STATE Render surround-view images $I_t \leftarrow \text{DecoupleGS}(S_{\text{ego}}^{(t)}, \mathbb{S}_{\text{bg}}^{(t)}, \mathcal{S}_{\text{bg}}, \{\mathcal{V}_k\})$
    
    \STATE \textbf{// 2. Ego-Vehicle Inference}
    \STATE Predict ego trajectory $\mathcal{T}_{\text{ego}} \leftarrow \pi_{\text{ego}}(I_t, \text{Command})$
    \STATE Compute next ego state $S_{\text{ego}}^{(t+1)}$ via kinematic bicycle model tracking $\mathcal{T}_{\text{ego}}$
    
    \STATE \textbf{// 3. Background Agents Interactive Response}
    \FOR{each background agent $k \in \mathbb{S}_{\text{bg}}^{(t)}$}
        \STATE Identify surrounding leaders/followers including the Ego vehicle $S_{\text{ego}}^{(t)}$
        \STATE Target lane $L_{\text{target}} \leftarrow \mathrm{MOBIL}(\text{Agent}_k)$
        \STATE Target acceleration $a_k \leftarrow \mathrm{IDM}(\text{Agent}_k, \text{Leader}_{L_{\text{target}}})$
        \STATE Compute next agent state $S_k^{(t+1)}$ via kinematic physical update
    \ENDFOR
    
    \STATE \textbf{// 4. Global State Synchronization \& Metrics}
    \STATE Synchronize global scene: update all 3DGS local-to-world transforms $\mathcal{T}(t+1)$
    \STATE Calculate $\text{minTTC}_t$ and update Route Completion ($\text{RC}$)
    \STATE $t \leftarrow t + \Delta t$
\ENDWHILE
\RETURN Calculated $\text{DS}$ and $\text{minTTC}$ over the episode
\end{algorithmic}
\end{algorithm}

\section{Additional Qualitative Results: Scene Decoupling and Editing}
\label{sec:additional_results}

In this final section, we present extended qualitative results that showcase the versatility of our DecoupleGS framework in scene manipulation. By fundamentally decoupling the static background from dynamic agents, our method natively supports advanced editing capabilities without requiring any retraining or per-scene network optimization.

\begin{figure}[tb!]
  \centering
  \includegraphics[width=0.85\linewidth]{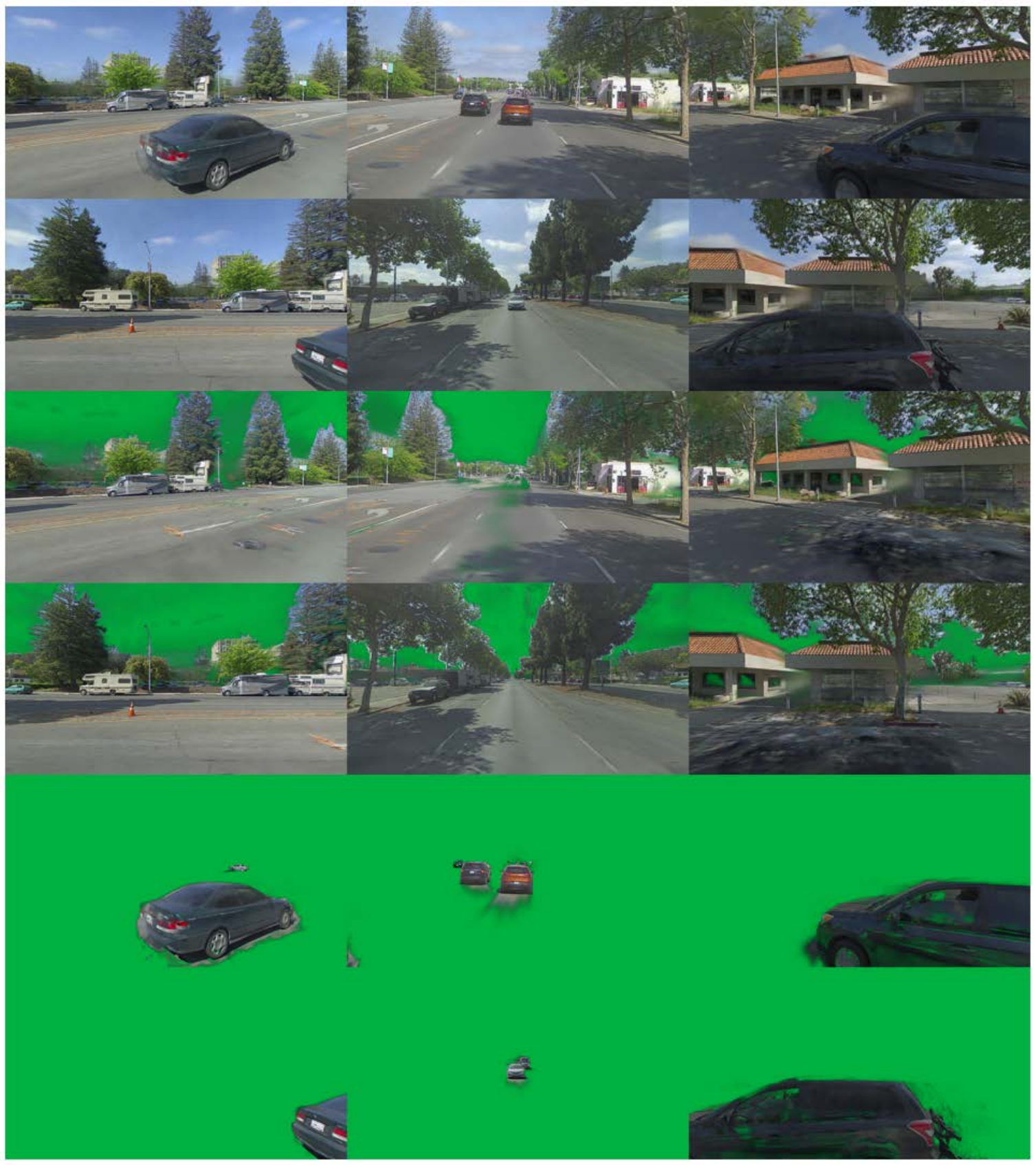}
  \caption{\textbf{Surround-View Scene Decoupling.} Top: Original reconstructed scene. Middle: Isolated static background with all dynamic agents perfectly removed. Bottom: Extracted dynamic foreground agents rendered independently. The decoupling is strictly consistent across the entire six-camera suite.}
  \label{fig:decoupling}
\end{figure}

\begin{figure}[tb!]
  \centering
  \includegraphics[width=0.85\linewidth]{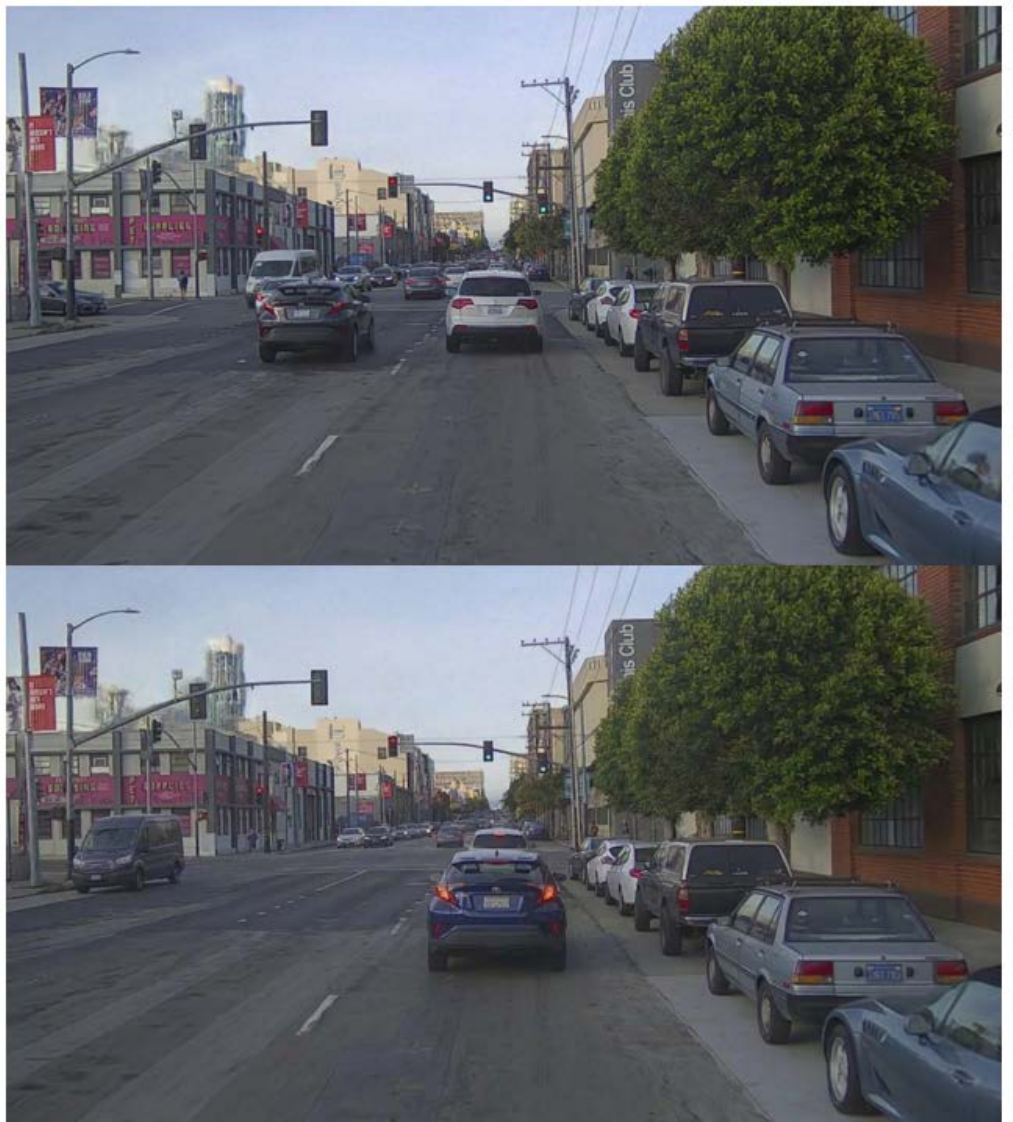}
  \caption{\textbf{Interactive Scene Editing.} Top: Original front-camera view containing native traffic. Bottom: Edited scene where vehicle appearances and spatial positions have been explicitly manipulated. The modified assets seamlessly integrate with the background's illumination and ground geometry.}
  \label{fig:editing}
\end{figure}

\subsection{Scene Decoupling in Surround-View}
\label{subsec:scene_decoupling}

The core advantage of our object-centric canonical representation is the ability to strictly isolate dynamic traffic participants from the persistent background across all Bird's-Eye-View (BEV) and sensor views. 

As demonstrated in Figure \ref{fig:decoupling}, our framework successfully decomposes a complex driving scene. The top row displays the original unedited reconstruction across the standard six-camera surround-view suite. The middle row isolates the static 3DGS background $\mathcal{S}_{\text{bg}}$, cleanly erasing all dynamic entities without leaving "floaters" or shadow entanglement artifacts typically seen in holistic neural representations. The bottom row visualizes the extracted dynamic canonical assets $\mathcal{V}_k$ rendered independently. Crucially, this decoupling maintains strict multi-view consistency and preserves the high-fidelity structural details of both the background infrastructure and the dynamic vehicles.

\subsection{Interactive Scene Editing}
\label{subsec:scene_editing}

Building upon the clean decoupling of the environment, DecoupleGS functions as a robust scenario generation engine that enables interactive and photorealistic scene editing. 

Figure \ref{fig:editing} illustrates this capability using the front-camera view. The original scene (top) contains the natively captured traffic flow. In the edited scene (bottom), we manipulate the dynamic assets by altering their spatial positions and substituting their appearances. Thanks to our map-guided registration and proxy-based relighting modules, the manipulated and newly inserted vehicles naturally inherit the geometric grounding and photometric illumination of the environment, achieving seamless integration. This confirms the framework's immense potential for generating diverse, adversarial closed-loop testing scenarios for E2E autonomous driving evaluation.

\end{document}